\documentclass[letterpaper,10pt]{IEEEtran}
\usepackage{graphicx,times,amsmath,amssymb,cite,hyperref,subfigure,url,multirow}
\usepackage[flushleft]{threeparttable}
\usepackage[lined,ruled,commentsnumbered]{algorithm2e}
 \allowdisplaybreaks

\begin{document}

\title{Feature Weighting Improves Pool-Based Sequential Active Learning for Regression}
\author{Dongrui~Wu,~\IEEEmembership{Fellow,~IEEE}
\thanks{Dongrui~Wu is with the Ministry of Education Key Laboratory of Image Processing and Intelligent Control, and Hubei Key Laboratory of Brain-inspired Intelligent Systems, School of Artificial Intelligence and Automation, Huazhong University of Science and Technology, Wuhan, 430074 China. Email: drwu09@gmail.com.}
\thanks{This research was supported by National Natural Science Foundation of China (62525305).}}

\markboth{IEEE TRANSACTIONS ON NEURAL NETWORKS AND LEARNING SYSTEMS}%
{Wu: }
\maketitle

\begin{abstract}
Pool-based sequential active learning for regression (ALR) optimally selects a small number of samples sequentially from a large pool of unlabeled samples to label, so that a more accurate regression model can be constructed under a given labeling budget. Representativeness and diversity, which involve computing the distances among different samples, are important considerations in ALR. However, previous ALR approaches do not incorporate the importance of different features in inter-sample distance computation, resulting in inaccurate distances and hence sub-optimal sample selection. This paper proposes four feature weighted single-task ALR approaches and three feature weighted multi-task ALR approaches, where the ridge regression coefficients trained from a small amount of previously labeled samples are used to weight the corresponding features in inter-sample distance computation. Extensive experiments showed that this intuitive and easy-to-implement enhancement almost always improves the performance of five existing ALR approaches, in both single-task and multi-task regression problems. The feature weighting strategy may also be easily extended to stream-based ALR, and classification algorithms.
\end{abstract}


\section{Introduction}

In many real-world machine learning problems, acquiring a large amount of unlabeled data is relatively easy, but obtaining their labels is challenging, as it may be:
\begin{enumerate}
\item \emph{Labor intensive}. For example, in affective computing~\cite{Picard1997,drwuPIEEE2023}, it is easy to collect affective input signals such as sound, images, videos, etc., but labelling each piece of such signals may require the aggregation of individual labels from tens or more evaluators, due to the uncertainty and subtleness of emotions: the VAM (\emph{Vera am Mittag} in German, \emph{Vera at Noon} in English) corpus~\cite{Grimm2008} used 6-17 evaluators for each speech utterance, the DEAP (Database for Emotion Analysis using Physiological Signals) dataset~\cite{Koelstra2012} used 14-16 assessors for each video clip, the SEED-IV (Shanghai Jiao Tong University Emotion EEG Dataset) dataset~\cite{Zheng2019a} used 44 evaluators for each film clip, the GAPED (Geneva Affective Picture Database) dataset~\cite{DanGlauser2011} used 60 college students for each picture, and the IADS-2 (International Affective Digitized Sound, 2nd Edition) dataset~\cite{Bradley2007} used over 110 assessors for each piece of sound.
\item \emph{Time consuming}. For example, in fracturing based enhanced oil recovery in the oil and gas industry~\cite{drwuSPE2009}, 180-day post-fracturing cumulative oil production is usually used as a performance measure. The fracturing parameters of an oil well, e.g., location, length of perforations, number of zones/holes, volumes of injected slurry/water/sand, etc., can be easily recorded; however, to get the groundtruth output (180-day post-fracturing cumulative oil production), one has to wait at least 180 days, which is very time consuming.
\item \emph{Expensive}. Usually one needs to pay to recruit tens or hundreds of evaluators for the affective sound/images/videos, so there is considerable cost to label each piece of affective input. Fracturing an oil well and observing its 180-day cumulative oil production is even more expensive.
\end{enumerate}
So, an important and realistic question is: \emph{how to achieve the best possible learning performance from a small number of labeled samples?}

Different strategies have been proposed for this purpose, e.g., active learning~\cite{Settles2012}, transfer learning~\cite{Yang2020a,drwuNTL2023}, self-supervised learning~\cite{Gui2024}, their combinations~\cite{Zhao2013,drwuSMC2017,drwuTAI2024,drwuDS3TL2024}, and so on. This paper considers pool-based sequential active learning, which optimally and sequentially selects a small number of samples from a large pool of unlabeled samples to label, so that a more accurate machine learning model can be constructed under a given (small) labeling budget. This is a very common active learning scenario. For example, in affective computing mentioned above, a large amount of affective input signals are usually collected offline before recruiting assessors for evaluation.

There have been many pool-based sequential active learning approaches proposed for classification problems~\cite{Settles2012,Li2025}, e.g., uncertainty sampling, expected model change maximization, query-by-committee, etc. However, pool-based sequential active learning for regression (ALR) problem is much less studied. As an evidence, the recent deep active learning survey~\cite{Li2025} published in 2025 includes 205 references, among which only one is on ALR. This paper focuses on this under-explored pool-based sequential ALR problem.

Previous research~\cite{drwuSAL2019} has proposed three important criteria to be considered in pool-based sequential ALR for selecting the next unlabeled sample to label:
\begin{enumerate}
\item \emph{Informativeness}, i.e., the selected sample must contain rich information, which could be measured by the disagreement among multiple learners~\cite{RayChaudhuri1995}, or expected model change~\cite{Cai2013}, etc.
\item \emph{Representativeness}, i.e., the selected sample should represent a large population, instead of being an outlier.
\item \emph{Diversity}, i.e., all selected samples together should scatter across the full input space, instead of occupying only a small local region.
\end{enumerate}

Based on these three criteria, several ALR approaches have been proposed, e.g., query-by-committee~\cite{RayChaudhuri1995}, greedy sampling~\cite{Yu2010}, expected model change maximization~\cite{Cai2013}, representativeness-diversity (RD)~\cite{drwuSAL2019}, greedy sampling in the input space (GSx)~\cite{drwuiGS2019}, greedy sampling in the output space~\cite{drwuiGS2019}, improved greedy sampling (iGS) in both input and output spaces~\cite{drwuiGS2019}, graph-based ALR (GALR)~\cite{Zhang2020}, and the multi-task iGS (MT-iGS)~\cite{drwuMTALR2022}. Many of them, e.g., RD, GSx, iGS, GALR and MT-iGS, need to compute the distances among different samples when calculating the representativeness and/or diversity measures. However, their distance computation does not consider the weights associated with the input features (which can be estimated from the small amount of labeled samples), resulting in inaccurate distances and hence sub-optimal performance.

This paper proposes a simple yet effective remedy, showing that by incorporating the estimated feature weights in inter-sample distance computation, ALR performance can be consistently improved. Our main contributions are:
\begin{enumerate}
\item We propose four feature weighted (FW) single-task ALR approaches, i.e., FW-RD, FW-GSx, FW-GALR and FW-iGS, to improve RD, GSx, GALR and iGS, respectively.
\item We propose three feature weighted multi-task (FW-MT) ALR approaches, i.e., FW-MT-GSx, FW-MT-GALR and FW-MT-iGS, to improve GSx, GALR (both GSx and GALR can be used in both single task ALR and multi-task ALR) and MT-iGS, respectively.
\item We perform extensive experiments to demonstrate that each of the seven feature weighted ALR approaches outperformed its corresponding unweighted version, and feature weighting benefits both linear and nonlinear prediction models.
\end{enumerate}

The remainder of this paper is organized as follows: Section~\ref{sect:existing} introduces five existing pool-based sequential ALR approaches, i.e., RD, GSx, iGS, GALR and MT-iGS, which need to compute the distances among different samples. Section~\ref{sect:FW} proposes seven feature weighted ALR approaches, FW-RD, FW-GSx, FW-iGS, FW-GALR, FW-MT-GSx, FW-MT-iGS and FW-MT-ALR. Sections~\ref{sect:experiment1} and \ref{sect:experiment2} perform experiments to demonstrate the effectiveness and robustness of the single-task and multi-task feature weighted ALR approaches, respectively. Finally, Section~\ref{sect:conclusions} draws conclusions and points out future research directions.

\section{Related Work} \label{sect:existing}

This section introduces five existing pool-based sequential ALR approaches, RD~\cite{drwuSAL2019}, GSx~\cite{drwuiGS2019}, iGS~\cite{drwuiGS2019}, GALR~\cite{Zhang2020} and MT-iGS~\cite{drwuMTALR2022}, which involve inter-sample distance computation in sample selection.

As in previous work~\cite{drwuSAL2019}, we assume the pool consists of $N$ $d$-dimensional samples $\{\mathbf{x}_n\}_{n=1}^N$, $\mathbf{x}_n\in \mathbb{R}^d$. Initially, none of the samples are labeled. An ALR algorithm first selects $d+1$ samples to label, so that a ridge regression model\footnote{Other (nonlinear) regression models, e.g., neural networks and regression trees, may also be constructed. However, when the number of labeled samples is very small, which is intrinsic in ALR, ridge regression achieves the most efficient, stable and generalizable performance. So, ridge regression is used in this paper.} can be properly initialized. It then selects a new sample to label in each iteration, until $M$ samples are labeled.

\subsection{RD} \label{sect:RD}

RD~\cite{drwuSAL2019} considers both representativeness and diversity, as its name suggests. It first selects simultaneously $d+1$ unlabeled samples from the pool to label, so that a ridge regression model can be initialized. Then, it selects one sample to label in each subsequent iteration, and updates the ridge regression model accordingly, until $M$ samples have been selected.

In initialization, RD performs $k$-means ($k=d+1$) clustering on the $N$ unlabeled samples, and then selects from each cluster (to achieve diversity) the sample closest to its centroid (to achieve representativeness) for labeling.

Assume $m$ ($m\ge d+1$) samples have been selected and labeled. To select the $(m+1)$-th sample to label, RD first performs $k$-means ($k=m+1$) clustering on all $N$ samples. Since there are only $m$ labeled samples but $m+1$ clusters, there must exist at least one cluster that does not contain any labeled samples. RD then identifies the largest such cluster (to achieve diversity), and selects the sample closest to its centroid (to achieve representativeness) for labeling. This process is repeated until $M$ samples are selected and labeled.

Algorithm~\ref{alg:RDAL} gives the pseudo-code of RD.

\begin{algorithm}[h]
\KwIn{$N$ unlabeled samples, $\{\mathbf{x}_n\}_{n=1}^N$, $\mathbf{x}_n\in \mathbb{R}^d$\;
\hspace*{9mm} $M$, the maximum number of samples to label.}
\KwOut{A ridge regression model $f(\mathbf{x})$.}
\tcp{Select the first $d+1$ samples to label}
Perform $k$-means ($k=d+1$) clustering on $\{\mathbf{x}_n\}_{n=1}^N$\;
Select from each cluster the sample closest to its centroid to label\;
\tcp{Iteratively select the next $M-d-1$ samples to label}
\For{$m=d+2,...,M$}{
Perform $k$-means ($k=m$) clustering on $\{\mathbf{x}_n\}_{n=1}^N$\;
Identify the largest cluster that does not contain any labeled samples\;
Select the sample closest to the cluster centroid to label\;}
Construct a ridge regression model $f(\mathbf{x})$ from the $M$ labeled samples.
\caption{The RD ALR algorithm~\cite{drwuSAL2019}.} \label{alg:RDAL}
\end{algorithm}

\subsection{GSx} \label{sect:GSx}

GSx~\cite{drwuiGS2019} considers the representativeness in first sample initialization, and then uses greedy sampling~\cite{Yu2010} to consider the diversity in the input space in subsequent sample selections. It selects the first sample as the one closest to the centroid of all $N$ samples, and the remaining $M-1$ samples iteratively.

Assume the first $m$ ($m\ge1$) samples have been selected, and they form a labeled set $L=\{(\mathbf{x}_i,y_i)\}_{i=1}^m$, where $y_i$ is the label for $\mathbf{x}_i$. The remaining $N-m$ unlabeled samples form an unlabeled set $U=\{\mathbf{x}_j\}_{j=1}^{N-m}$. GSx needs to select a sample from $U$ to label next. For each $\mathbf{x}_j\in U$, GSx computes first its distance to each $\mathbf{x}_i\in L$:
\begin{align}
d_{ij}^{\mathbf{x}}=||\mathbf{x}_i-\mathbf{x}_j||,\quad \forall \mathbf{x}_i\in L, \forall \mathbf{x}_j\in U, \label{eq:dnmx}
\end{align}
then $d_j^{\mathbf{x}}$, the shortest distance from $\mathbf{x}_j$ to all $m$ labeled samples in $L$:
\begin{align}
d_j^{\mathbf{x}}=\min_i d_{ij}^{\mathbf{x}},\quad \forall \mathbf{x}_j\in U \label{eq:dnx}
\end{align}
and selects the sample with the maximum $d_j^{\mathbf{x}}$ to label, to achieve diversity in the input space.

Algorithm~\ref{alg:GSx} gives the pseudo-code of GSx.

\begin{algorithm}[h]
\KwIn{$N$ unlabeled samples, $\{\mathbf{x}_n\}_{n=1}^N$, $\mathbf{x}_n\in \mathbb{R}^d$\;
\hspace*{10mm} $M$, the maximum number of samples to label.}
\KwOut{A ridge regression model $f(\mathbf{x})$.}
\tcp{Select the first sample to label}
Set $U=\{\mathbf{x}_n\}_{n=1}^N$, and $L=\emptyset$\;
Identify $\mathbf{x}_*$, the sample closest to the centroid of $U$, to label\;
Move $\mathbf{x}_*$ from $U$ to $L$\;
\tcp{Iteratively select the next $M-1$ samples to label}
\For{$m=1,...,M-1$}{
Compute $d_j^{\mathbf{x}}$ in (\ref{eq:dnx}) for each $\mathbf{x}_j\in U$\;
Select $\mathbf{x}_*$ with the largest $d_j^{\mathbf{x}}$ to label\;
Move $\mathbf{x}_*$ from $U$ to $L$\;}
Construct a ridge regression model $f(\mathbf{x})$ from the $M$ labeled samples in $L$.
\caption{The GSx ALR algorithm~\cite{drwuiGS2019}.} \label{alg:GSx}
\end{algorithm}

Another very popular active learning strategy is core-set~\cite{Sener2018}, whose sequential sample selection strategy is identical to greedy sampling~\cite{Yu2010}. So, the iterative sample selection strategy of GSx and core-set are also identical; however, GSx and core-set have two differences: 1) GSx selects the first sample by considering the representativeness (closest to the centroid of all unlabeled samples), whereas core-set selects the first sample (or first few samples) randomly; and, 2) GSx was original proposed for regression problems, whereas core-set was originally proposed for classification problems, although both can be used in both regression and classification.

In summary, GSx can be viewed as an enhanced version of core-set for ALR; so, core-set is not separately introduced in this paper.

\subsection{iGS} \label{sect:iGS}

iGS~\cite{drwuiGS2019} considers the diversity in both the input space and the output space.

iGS uses GSx to select the first $d+1$ samples to label.

Let $L=\{(\mathbf{x}_i,y_i)\}_{i=1}^m$ be the set of $m$ ($m\ge d+1$) samples that have already been labeled, and $U=\{\mathbf{x}_j\}_{j=1}^{N-m}$ be the set of $N-m$ unlabeled samples. iGS constructs a ridge regression model $f(\mathbf{x})$ from $L$. Then, for each $\mathbf{x}_j\in U$, iGS computes $d_{ij}^{\mathbf{x}}$ in (\ref{eq:dnmx}), and $d_{ij}^y$ below:
\begin{align}
d_{ij}^y=|f(\mathbf{x}_j)-y_i|,\quad \forall \mathbf{x}_i\in L, \forall \mathbf{x}_j\in U, \label{eq:dnmy}
\end{align}
and then $d_j^{\mathbf{x}y}$:
\begin{align}
d_j^{\mathbf{x}y}=\min_i d_{ij}^{\mathbf{x}}\cdot d_{ij}^y,\quad \forall \mathbf{x}_j\in U. \label{eq:dnxy}
\end{align}
It next selects the sample with the maximum $d_j^{\mathbf{x}y}$ to label, to achieve diversity in simultaneously the input space and the output space.

Algorithm~\ref{alg:iGS} gives the pseudo-code of iGS.

\begin{algorithm}[h]
\KwIn{$N$ unlabeled samples, $\{\mathbf{x}_n\}_{n=1}^N$, $\mathbf{x}_n\in \mathbb{R}^d$\;
\hspace*{10mm} $M$, the maximum number of samples to label.}
\KwOut{A ridge regression model $f(\mathbf{x})$.}
\tcp{Select the first sample to label, using GSx}
Set $U=\{\mathbf{x}_n\}_{n=1}^N$, and $L=\emptyset$\;
Identify $\mathbf{x}_*$, the sample closest to the centroid of $U$, to label\;
Move $\mathbf{x}_*$ from $U$ to $L$\;
\tcp{Iteratively select the next $d$ samples to label, using GSx}
\For{$m=2,...,d+1$}{
Compute $d_j^\mathbf{x}$ in (\ref{eq:dnx}) for each $\mathbf{x}_j\in U$\;
Select $\mathbf{x}_*$ with the largest $d_j^{\mathbf{x}}$ to label\;
Move $\mathbf{x}_*$ from $U$ to $L$\;}
Construct a ridge regression model $f(\mathbf{x})$ from the $d+1$ labeled samples in $L$\;
\tcp{Iteratively select the next $M-d-1$ samples to label}
\For{$m=d+2,...,M$}{
Compute $d_j^{\mathbf{x}y}$ in (\ref{eq:dnxy}) for each $\mathbf{x}_j\in U$\;
Select $\mathbf{x}_*$ with the largest $d_j^{\mathbf{x}y}$ to label\;
Move $\mathbf{x}_*$ from $U$ to $L$\;
Update $f(\mathbf{x})$ using the $m$ labeled samples in $L$.}
\caption{The iGS ALR algorithm~\cite{drwuiGS2019}.} \label{alg:iGS}
\end{algorithm}

\subsection{Graph-based ALR (GALR)}  \label{sect:GALR}

Zhang \emph{et al.} ~\cite{Zhang2020} proposed GALR, using only feature space properties of the unlabeled samples. In each iteration, it selects the sample that will maximally reduce the uncertainty to label, i.e., it focuses only on the informativeness. However, its final formulation is almost identical to GSx, except that $\mathcal{L}_1$ distance instead of $\mathcal{L}_2$ distance is used.

More specifically, GALR replaces (\ref{eq:dnmx}) in GSx by
\begin{align}
d_{ij}^{\mathbf{x}}=|\mathbf{x}_i-\mathbf{x}_j|,\quad \forall \mathbf{x}_i\in L, \forall \mathbf{x}_j\in U. \label{eq:dnmxGALR}
\end{align}
Due to the high similarity between GALR and GSx, and the page limit, the pseudo-code of GALR is omitted.

%

\subsection{MT-iGS} \label{sect:MT-iGS}

MT-iGS~\cite{drwuMTALR2022} extends iGS from single-task ALR to multi-task ALR, by considering also simultaneously the diversity in the input and output spaces. More specifically, it considers the scenario that $T$ different regression tasks share the same input, e.g., three separate tasks to estimate the valence, arousal and dominance of emotions from the same speech utterance. Its goal is to optimally select some unlabeled samples to label, so that all $T$ tasks can be benefited simultaneously. A typical performance measure is the average performance of the $T$ tasks.

MT-iGS~\cite{drwuMTALR2022} uses GSx to select the first $d+1$ samples to label (note that each sample has $T$ labels, each corresponding to a different task). It then selects the remaining $M-d-1$ samples to achieve diversity in both the input and output spaces.

Let $L=\{(\mathbf{x}_i,y_{i,1},...,y_{i,T})\}_{i=1}^m$ be the set of $m$ ($m\ge d+1$) samples that have already been labeled, where $y_{i,t}$ is the label for the $t$-th task. and $U=\{\mathbf{x}_j\}_{j=1}^{N-m}$ be the set of $N-m$ unlabeled samples. $T$ ridge regression models $\{f_t(\mathbf{x})\}_{t=1}^T$ can be constructed from $L$, each for a different task. For each of the remaining $N-m$ unlabeled samples in $U$, MT-iGS computes $d_{ij}^{\mathbf{x}}$ in (\ref{eq:dnmx}) and $d_{ij,t}^\mathbf{y}$ below:
\begin{align}
d_{ij,t}^\mathbf{y}=||f_t(\mathbf{x}_j)-y_{i,t}|| \quad \forall \mathbf{x}_i\in L, \forall \mathbf{x}_j\in U  \label{eq:dnmyp}
\end{align}
and then $d_j^\mathbf{xy}$:
\begin{align}
d_j^\mathbf{xy}=\min_i d_{ij}^{\mathbf{x}}\cdot \prod_{t=1}^T d_{ij,t}^\mathbf{y},\quad \forall \mathbf{x}_j\in U. \label{eq:dnxyp}
\end{align}
It next selects the sample with the maximum $d_j^\mathbf{xy}$ to label, to achieve diversity in simultaneously the input space and the output spaces for all tasks.

Algorithm~\ref{alg:MT-iGS} gives the pseudo-code of MT-iGS.

\begin{algorithm}[h] 
\KwIn{$N$ unlabeled samples, $\{\mathbf{x}_n\}_{n=1}^N$, $\mathbf{x}_n\in \mathbb{R}^d$\;
\hspace*{10mm} $M$, the maximum number of samples to label.}
\KwOut{$T$ ridge regression models $\{f_t(\mathbf{x})\}_{t=1}^T$.}
\tcp{Select the first sample to label, using GSx}
Set $U=\{\mathbf{x}_n\}_{n=1}^N$, and $L=\emptyset$\;
Identify $\mathbf{x}_*$, the sample closest to the centroid of $U$\;
Move $\mathbf{x}_*$ from $U$ to $L$\;
\tcp{Iteratively select the next $d$ samples to label, using GSx}
\For{$m=2,...,d+1$}{
Compute $d_j^\mathbf{x}$ in (\ref{eq:dnx}) for each $\mathbf{x}_j\in U$\;
Select $\mathbf{x}_*$ with the largest $d_j^{\mathbf{x}}$ to label\;
Move $\mathbf{x}_*$ from $U$ to $L$\;}
Construct $T$ regression models $\{f_t(\mathbf{x})\}_{t=1}^T$ from $L$, one for each task\;
\tcp{Iteratively select the next $M-d-1$ samples to label}
\For{$m=d+2,...,M$}{
Compute $d_j^\mathbf{xy}$ in (\ref{eq:dnxyp}) for each $\mathbf{x}_j\in U$\;
Select $\mathbf{x}_*$ with the largest $d_j^{\mathbf{xy}}$ to label\;
Move $\mathbf{x}_*$ from $U$ to $L$\;
Update $\{f_t(\mathbf{x})\}_{t=1}^T$ using the $m$ labeled samples in $L$\;}
\caption{The MT-iGS multi-task ALR algorithm~\cite{drwuMTALR2022}.} \label{alg:MT-iGS}
\end{algorithm}

\section{Feature Weighted ALR} \label{sect:FW}

The five ALR approaches introduced in the previous section all involve computing the distances among the samples, e.g., in $k$-means clustering and in (\ref{eq:dnmx}) and (\ref{eq:dnmxGALR}); however, they do not take the importance of different features into consideration. As a result, the distances may be problematic:
\begin{enumerate}
\item Within a single feature, its value changes with different units, and hence the distances between two samples also changes with units, which should not be the case. For example, in the UCI autoMPG dataset used in our experiments, the weight of a car can be measured in pounds or kilograms or tons (e.g., 3504 pounds = 1589.39 kilograms = 1.58939 tons) , which does not change its characteristics at all; however, using weight 3504/1589.39/1.58939 in computing the distances among different cars (when other feature values keep the same) gives significantly different results, hence dramatically different representativeness and/or diversity values, and finally different sample selections.
\item Among different features, those with large magnitudes dominate the distance computation, whereas they may not be the most important features.
\item Some features may be redundant or irrelevant in real-world machine learning problems. Blindly using all features in distance computation is misleading, resulting in wrong representativeness and/or diversity values, and hence suboptimal sample selections.
\end{enumerate}

So, we argue that the feature weights, which can be estimated from ridge regression or tree models, should be used to more accurately computing the distances, and hence to improve sample selection:
\begin{enumerate}
\item Within a single feature, when feature weight is considered in distance computation, i.e., the value of the feature is replaced by (feature weight $\times$ feature value), the distances no longer change with the units of the features, and hence are more reasonable.
\item Among different features, when (feature weight $\times$ feature value) is used in inter-sample distance computation, the contributions of unimportant features with large magnitudes is suppressed by their small feature weights, avoiding misleading distances.
\item The contributions of redundant or irrelevant features to the distances may also be automatically adjusted by the feature weights, minimizing their negative impacts to sample selection.
\end{enumerate}

This section introduces our proposed seven feature weighted ALR algorithms.

\subsection{FW-RD} \label{sect:fwRD}

FW-RD also considers simultaneously diversity and representativeness. As in the original RD algorithm in Section~\ref{sect:RD}, FW-RD first selects $d+1$ unlabeled samples from the pool to label, so that a ridge regression model can be initialized. Then, it selects one sample to label in each iteration, and updates the ridge regression model accordingly, until $M$ samples have been selected.

In initialization, FW-RD performs $k$-means ($k=d+1$) clustering on the $N$ unlabeled samples, and then selects from each cluster (to achieve diversity) the sample closest to its centroid (to achieve representativeness) for labeling.

Assume $m$ ($m\ge d+1$) samples have been selected and labeled. A ridge regression model can then be constructed from them. Let the regression coefficient vector be $\mathbf{w}\in\mathbb{R}^d$. To select the $(m+1)$-th sample to label, FW-RD first multiplies each dimension of $\mathbf{x}_n$ by the corresponding regression coefficient in $\mathbf{w}$, i.e.,
\begin{align}
\mathbf{x}_n^\mathrm{w}=\mathbf{w}\odot \mathbf{x}_n,\quad n=1,...,N, \label{eq:xnw}
\end{align}
where $\odot$ is element-wise multiplication. FW-RD then performs $k$-means ($k=m+1$) clustering on $\{\mathbf{x}_n^\mathrm{w}\}_{n=1}^N$. Since there are only $m$ labeled samples but $m+1$ clusters, there must exist at least one cluster that does not contain any labeled samples. FW-RD next identifies the largest such cluster (to achieve diversity), and selects the sample closest to its centroid (to achieve representativeness) for labeling. This process is repeated until $M$ samples are selected and labeled.

Algorithm~\ref{alg:fwRDAL} gives the pseudo-code of FW-RD. It is very similar to the pseudo-code of RD in Algorithm~\ref{alg:RDAL}, except that the feature weighted samples are used in $k$-means clustering when $m>d+1$.

\begin{algorithm}[h]
\KwIn{$N$ unlabeled samples, $\{\mathbf{x}_n\}_{n=1}^N$, $\mathbf{x}_n\in \mathbb{R}^d$\;
\hspace*{9mm} $M$, the maximum number of samples to label.}
\KwOut{A ridge regression model $f(\mathbf{x})$.}
\tcp{Select the first $d+1$ samples to label}
Perform $k$-means ($k=d+1$) clustering on $\{\mathbf{x}_n\}_{n=1}^N$\;
Select from each cluster the sample closest to its centroid to label\;
Construct a ridge regression model $f(\mathbf{x})$ from the $d+1$ labeled samples\;
\tcp{Iteratively select the next $M-d-1$ samples to label}
\For{$m=d+2,...,M$}{
Let $\mathbf{w}$ be the regression coefficient vector in $f(\mathbf{x})$\;
Compute $\mathbf{x}_n^\mathrm{w}$ in (\ref{eq:xnw})\;
Perform $k$-means ($k=m$) clustering on $\{\mathbf{x}_n^\mathrm{w}\}_{n=1}^N$\;
Identify the largest cluster that does not contain any labeled samples\;
Select the sample closest to the cluster centroid to label\;
Update $f(\mathbf{x})$ using the $m$ labeled samples\;}
\caption{The FW-RD ALR algorithm.} \label{alg:fwRDAL}
\end{algorithm}

\subsection{FW-GSx} \label{sect:fwGSx}

Similar to GSx, FW-GSx considers mainly the diversity in the input space.

During initialization, FW-GSx uses GSx to select the first $d+1$ samples for labeling.

Assume the first $m$ ($m\ge d+1$) samples have been selected, and they form a labeled set $L=\{(\mathbf{x}_i,y_i)\}_{i=1}^m$. A ridge regression model can then be constructed from $L$. The remaining $N-m$ unlabeled samples form an unlabeled set $U=\{\mathbf{x}_j\}_{j=1}^{N-m}$. FW-GSx first computes $\mathbf{x}_n^\mathrm{w}$ by (\ref{eq:xnw}). It then needs to select a sample from $U$ to label next. For each $\mathbf{x}_j\in U$, FW-GSx computes first its feature weighted distance to each $\mathbf{x}_i\in L$:
\begin{align}
d_{ij}^{\mathbf{x}^\mathrm{w}}=||\mathbf{x}_i^\mathrm{w}-\mathbf{x}_j^\mathrm{w}||,\quad \forall \mathbf{x}_i\in L, \forall \mathbf{x}_j\in U, \label{eq:fwdnmx}
\end{align}
then $d_j^{\mathbf{x}}$, the shortest feature weighted distance from $\mathbf{x}_j$ to all $m$ labeled samples in $L$:
\begin{align}
d_j^{\mathbf{x}^\mathrm{w}}=\min_i d_{ij}^{\mathbf{x}^\mathrm{w}},\quad \forall \mathbf{x}_j\in U. \label{eq:fwdnx}
\end{align}
FW-GSx next selects the sample with the maximum $d_j^{\mathbf{x}^\mathrm{w}}$ to label, to achieve diversity in the input space.

Algorithm~\ref{alg:fwGSx} gives the pseudo-code of FW-GSx. It is very similar to the pseudo-code of GSx in Algorithm~\ref{alg:GSx}, except that the feature weights are used in computing the inter-sample distances when $m>d+1$.

\begin{algorithm}[htpb]
\KwIn{$N$ unlabeled samples, $\{\mathbf{x}_n\}_{n=1}^N$, $\mathbf{x}_n\in \mathbb{R}^d$\;
\hspace*{10mm} $M$, the maximum number of samples to label.}
\KwOut{A ridge regression model $f(\mathbf{x})$.}
\tcp{Select the first sample to label, using GSx}
Set $U=\{\mathbf{x}_n\}_{n=1}^N$, and $L=\emptyset$\;
Identify $\mathbf{x}_*$, the sample closest to the centroid of $U$, to label\;
Move $\mathbf{x}_*$ from $U$ to $L$\;
\tcp{Iteratively select the next $d$ samples to label, using GSx}
\For{$m=2,...,d+1$}{
Compute $d_j^\mathbf{x}$ in (\ref{eq:dnx}) for each $\mathbf{x}_j\in U$\;
Select $\mathbf{x}_*$ with the largest $d_j^{\mathbf{x}}$ to label\;
Move $\mathbf{x}_*$ from $U$ to $L$\;}
Construct a ridge regression model $f(\mathbf{x})$ from the $d+1$ labeled samples\;
\tcp{Iteratively select the next $M-d-1$ samples to label}
\For{$m=d+2,...,M$}{
Let $\mathbf{w}$ be the regression coefficient vector in $f(\mathbf{x})$\;
Compute $\mathbf{x}_n^\mathrm{w}$ in (\ref{eq:xnw})\;
Compute $d_j^{\mathbf{x}^\mathrm{w}}$ in (\ref{eq:fwdnx}) for all $\mathbf{x}_j\in U$\;
Select $\mathbf{x}_*$ with the largest $d_j^{\mathbf{x}^\mathrm{w}}$ to label\;
Move $\mathbf{x}_*$ from $U$ to $L$\;
Update $f(\mathbf{x})$ using the $m$ labeled samples in $L$\;}
\caption{The FW-GSx ALR algorithm.} \label{alg:fwGSx}
\end{algorithm}

As pointed out in Section~\ref{sect:GSx}, the iterative sample selection strategy of GSx and core-set are identical; hence, feature weighted core-set can be easily modified from FW-GSx, by selecting the first sample (or the first few samples) randomly, instead of by GSx as in FW-GSx.

\subsection{FW-iGS} \label{sect:fwiGS}

Similar to iGS, FW-iGS also considers simultaneously the diversity in the input space and output space.

During initialization, FW-iGS uses GSx to select the first $d+1$ samples to label.

Let $L=\{(\mathbf{x}_i,y_i)\}_{i=1}^m$ be the set of $m$ ($m\ge d+1$) samples that have already been labeled, and $U=\{\mathbf{x}_j\}_{j=1}^{N-m}$ be the set of $N-m$ unlabeled samples. FW-iGS constructs a ridge regression model $f(\mathbf{x})$ from $L$, and computes $\mathbf{x}_n^\mathrm{w}$ by (\ref{eq:xnw}). Next, for each $\mathbf{x}_j\in U$, FW-iGS computes $d_{ij}^{\mathbf{x}^\mathrm{w}}$ in (\ref{eq:fwdnmx}), $d_{ij}^y$ in (\ref{eq:dnmy}), and then the following $d_j^{\mathbf{x}^\mathrm{w}y}$:
\begin{align}
d_j^{\mathbf{x}^\mathrm{w}y}=\min_i d_{ij}^{\mathbf{x}^\mathrm{w}}\cdot d_{ij}^y,\quad \forall \mathbf{x}_j\in U. \label{eq:fwdnxy}
\end{align}
FW-iGS next selects the sample with the maximum $d_j^{\mathbf{x}^\mathrm{w}y}$ to label, to achieve diversity in both the input space and the output space.

Algorithm~\ref{alg:fwiGS} gives the pseudo-code of FW-iGS. It is very similar to the pseudo-code of iGS in Algorithm~\ref{alg:iGS}, except that the feature weights are used in computing the inter-sample distances when $m>d+1$.

\begin{algorithm}[htpb]
\KwIn{$N$ unlabeled samples, $\{\mathbf{x}_n\}_{n=1}^N$, $\mathbf{x}_n\in \mathbb{R}^d$\;
\hspace*{10mm} $M$, the maximum number of samples to label.}
\KwOut{A ridge regression model $f(\mathbf{x})$.}
\tcp{Select the first sample to label, using GSx}
Set $U=\{\mathbf{x}_n\}_{n=1}^N$, and $L=\emptyset$\;
Identify $\mathbf{x}_*$, the sample closest to the centroid of $U$, to label\;
Move $\mathbf{x}_*$ from $U$ to $L$\;
\tcp{Iteratively select the next $d$ samples to label, using GSx}
\For{$m=2,...,d+1$}{
Compute $d_j^\mathbf{x}$ in (\ref{eq:dnx}) for all $\mathbf{x}_j\in U$\;
Select $\mathbf{x}_*$ with the largest $d_j^{\mathbf{x}}$ to label\;
Move $\mathbf{x}_*$ from $U$ to $L$\;}
Construct a ridge regression model $f(\mathbf{x})$ from the $d+1$ labeled samples in $L$\;
\tcp{Iteratively select the next $M-d-1$ samples to label}
\For{$m=d+2,...,M$}{
Let $\mathbf{w}$ be the regression coefficient vector in $f(\mathbf{x})$\;
Compute $\mathbf{x}_n^\mathrm{w}$ in (\ref{eq:xnw})\;
Compute $d_j^{\mathbf{x}^\mathrm{w}y}$ in (\ref{eq:fwdnxy}) for each $\mathbf{x}_j\in U$\;
Select $\mathbf{x}_*$ with the largest $d_j^{\mathbf{x}^\mathrm{w}y}$ to label\;
Move $\mathbf{x}_*$ from $U$ to $L$\;
Update $f(\mathbf{x})$ using the $m$ labeled samples in $L$\;}
\caption{The FW-iGS ALR algorithm.} \label{alg:fwiGS}
\end{algorithm}

\subsection{FW-GALR} \label{sect:fwGALR}

Since GALR~\cite{Zhang2020} is almost identical to GSx, FW-GALR is also almost identical to FW-GSx, except that $d_{ij}^{\mathbf{x}^\mathrm{w}}$ in (\ref{eq:fwdnmx}) of FW-GSx is replaced by:
\begin{align}
d_{ij}^{\mathbf{x}^\mathrm{w}}=|\mathbf{x}_i^\mathrm{w}-\mathbf{x}_j^\mathrm{w}|,\quad \forall \mathbf{x}_i\in L, \forall \mathbf{x}_j\in U, \label{eq:fwdnmxGALR}
\end{align}
i.e., feature weighted $\mathcal{L}_1$ distance, instead of $\mathcal{L}_2$ distance, is used.

Due to the high similarity between FW-GALR and FW-GSx, and the page limit, the pseudo-code of FW-GALR is omitted.

%

\subsection{FW-MT-GSx} \label{sect:fwMTGSx}

When the feature weights are not considered, GSx can be used for both single-task ALR and multi-task ALR. However, when incorporating the feature weights, a dedicated FW-MT-GSx has to be developed for multi-task ALR, because each task has different ridge regression coefficients (feature weights). Similar to GSx, FW-MT-GSx mainly considers the diversity in the input space.

During initialization, FW-MT-GSx uses GSx to select the first $d+1$ samples for labeling  (note that each sample has $T$ labels, each corresponding to a different task).

Assume the first $m$ ($m\ge d+1$) samples have been selected, and they form a labeled set $L=\{(\mathbf{x}_i,y_i)\}_{i=1}^m$. A ridge regression model can then be constructed from $L$. The remaining $N-m$ unlabeled samples form an unlabeled set $U=\{\mathbf{x}_j\}_{j=1}^{N-m}$. $T$ ridge regression models $\{f_t(\mathbf{x})\}_{t=1}^T$ can be constructed from $L$, one for each task.

Let $\mathbf{w}_t\in\mathbb{R}^d$ ($t=1,...,T$) be the regression coefficient vector for the $t$-th task. For the $t$-th task, FW-MT-GSx first multiplies each dimension of $\mathbf{x}_n$ by the corresponding regression coefficient in $\mathbf{w}_t$, i.e.,
\begin{align}
\mathbf{x}_n^{\mathrm{w}_t}=\mathbf{w}_t\odot \mathbf{x}_n,\quad \forall\mathbf{x}_n\in L\cup U;\ t=1,...,T, \label{eq:xnwMT}
\end{align}
where $\odot$ is element-wise multiplication. FW-MT-GSx then needs to select a sample from $U$ to label. For each $\mathbf{x}_j\in U$, FW-MT-GSx computes first its product of feature weighted distances to each $\mathbf{x}_i\in L$:
\begin{align}
d_{ij}^{\mathbf{x}^\mathrm{w}}=\prod_{t=1}^T\left\|\mathbf{x}_i^{\mathrm{w}_t}-\mathbf{x}_j^{\mathrm{w}_t}\right\|,\quad \forall \mathbf{x}_i\in L, \forall \mathbf{x}_j\in U, \label{eq:fwmtdnmx}
\end{align}
then $d_j^{\mathbf{x}}$, the shortest feature weighted distance from $\mathbf{x}_j$ to all $m$ labeled samples in $L$:
\begin{align}
d_j^{\mathbf{x}^\mathrm{w}}=\min_i d_{ij}^{\mathbf{x}^\mathrm{w}},\quad \forall \mathbf{x}_j\in U. \label{eq:fwmtdnx}
\end{align}
FW-MT-GSx next selects the sample with the maximum $d_j^{\mathbf{x}^\mathrm{w}}$ to label, to achieve diversity in the input space.

Algorithm~\ref{alg:fwMTGSx} gives the pseudo-code of FW-MT-GSx.

\begin{algorithm}[htpb]
\KwIn{$N$ unlabeled samples, $\{\mathbf{x}_n\}_{n=1}^N$, $\mathbf{x}_n\in \mathbb{R}^d$\;
\hspace*{10mm} $M$, the maximum number of samples to label.}
\KwOut{A ridge regression model $f(\mathbf{x})$.}
\tcp{Select the first sample to label, using GSx}
Set $U=\{\mathbf{x}_n\}_{n=1}^N$, and $L=\emptyset$\;
Identify $\mathbf{x}_*$, the sample closest to the centroid of $U$, to label\;
Move $\mathbf{x}_*$ from $U$ to $L$\;
\tcp{Iteratively select the next $d$ samples to label, using GSx}
\For{$m=2,...,d+1$}{
Compute $d_j^\mathbf{x}$ in (\ref{eq:dnx}) for all $\mathbf{x}_j\in U$\;
Select $\mathbf{x}_*$ with the largest $d_j^{\mathbf{x}}$ to label\;
Move $\mathbf{x}_*$ from $U$ to $L$\;}
\tcp{Iteratively select the next $M-d-1$ samples to label}
\For{$m=d+2,...,M$}{
\For{$t=1,...,T$}{
Construct a ridge regression model $f_t(\mathbf{x})$ from the $m-1$ labeled samples in $L$, for the $t$-th task\;
Let $\mathbf{w}_t$ be the regression coefficient vector\;
Compute $\mathbf{x}_n^{\mathrm{w}_t}$ in (\ref{eq:xnwMT})\;}
Compute $d_j^{\mathbf{x}^\mathrm{w}}$ in (\ref{eq:fwmtdnx}) for each $\mathbf{x}_j\in U$\;
Select $\mathbf{x}_*$ with the largest $d_j^{\mathbf{x}^\mathrm{w}}$ to label\;
Move $\mathbf{x}_*$ from $U$ to $L$\;}
\For{$t=1,...,T$}{
Update $f_t(\mathbf{x})$ using the $M$ labeled samples in $L$\;}
\caption{The FW-MT-GSx ALR algorithm.} \label{alg:fwMTGSx}
\end{algorithm}

\subsection{FW-MT-iGS} \label{sect:fwMT-iGS}

Similar to MT-iGS, FW-MT-iGS considers simultaneously the diversity in the input and output spaces.

During initialization, FW-MT-iGS uses GSx to select the first $d+1$ samples to label (note that each sample has $T$ labels, each corresponding to a different task). It then selects the remaining $M-d-1$ samples to achieve diversity in both the input and output spaces.

Let $L=\{(\mathbf{x}_i,y_{i,1},...,y_{i,T})\}_{i=1}^m$ be the set of $m$ ($m\ge d+1$) samples that have already been labeled, and $U=\{\mathbf{x}_j\}_{j=1}^{N-m}$ be the set of $N-m$ unlabeled samples. $T$ ridge regression models $\{f_t(\mathbf{x})\}_{t=1}^T$ can be constructed from $L$, one for each task. Let $\mathbf{w}_t$ be the regression coefficient vector for the $t$-th task. For the $t$-th task, FW-MT-iGS first computes $\mathbf{x}_n^{\mathrm{w}_t}$ by (\ref{eq:xnwMT}). It then needs to select a sample from $U$ to label. For each $\mathbf{x}_j\in U$, FW-MT-iGS computes first its product of feature weighted distances $d_{ij}^{\mathbf{x}^\mathrm{w}}$ to each $\mathbf{x}_i\in L$ using (\ref{eq:fwmtdnx}), and $d_{ij}^\mathbf{y}$ below:
\begin{align}
d_{ij}^\mathbf{y}=\prod_{t=1}^T|f_t(\mathbf{x}_j)-y_{i,t}|,\quad \forall \mathbf{x}_i\in L, \forall \mathbf{x}_j\in U, \label{eq:mtdnmy}
\end{align}
and then $d_j^\mathbf{xy}$:
\begin{align}
d_j^\mathbf{xy}=\min_i d_{ij}^{\mathbf{x}^\mathrm{w}}\cdot d_{ij}^\mathbf{y},\quad \forall \mathbf{x}_j\in U. \label{eq:fwdnxyp}
\end{align}
FW-MT-IGS next selects the sample with the maximum $d_j^\mathbf{xy}$ to label, to simultaneously achieve the diversity in the input and output spaces.

Algorithm~\ref{alg:fwMT-iGS} gives the pseudo-code of FW-MT-iGS. It is very similar to the pseudo-code of MT-iGS in Algorithm~\ref{alg:MT-iGS}, except that the feature weights are used in computing the inter-sample distances when $m>d+1$.

\begin{algorithm}[htpb] 
\KwIn{$N$ unlabeled samples, $\{\mathbf{x}_n\}_{n=1}^N$, $\mathbf{x}_n\in \mathbb{R}^d$\;
\hspace*{10mm} $M$, the maximum number of samples to label.}
\KwOut{$T$ ridge regression models $\{f_t(\mathbf{x})\}_{t=1}^T$.}
\tcp{Select the first sample to label, using GSx}
Set $U=\{\mathbf{x}_n\}_{n=1}^N$, and $L=\emptyset$\;
Identify $\mathbf{x}_*$, the sample closest to the centroid of $U$, to label\;
Move $\mathbf{x}_*$ from $U$ to $L$\;
\tcp{Iteratively select the next $d$ samples to label, using GSx}
\For{$m=2,...,d+1$}{
Compute $d_j^\mathbf{x}$ in (\ref{eq:dnx}) for each $\mathbf{x}_j\in U$\;
Select $\mathbf{x}_*$ with the largest $d_j^{\mathbf{x}}$ to label\;
Move $\mathbf{x}_*$ from $U$ to $L$\;}
\tcp{Iteratively select the next $M-d-1$ samples to label}
\For{$m=d+2,...,M$}{
\For{$t=1,...,T$}{
Construct a ridge regression model $f_t(\mathbf{x})$ from the $m-1$ labeled samples in $L$, for the $t$-th task\;
Let $\mathbf{w}_t$ be the regression coefficient vector\;
Compute $\mathbf{x}_n^{\mathrm{w}_t}$ in (\ref{eq:xnwMT})\;}
Compute $d_j^\mathbf{xy}$ in (\ref{eq:fwdnxyp}) for each $\mathbf{x}_j\in U$\;
Select $\mathbf{x}_*$ with the largest $d_j^{\mathbf{xy}}$ to label\;
Move $\mathbf{x}_*$ from $U$ to $L$\;}
\For{$t=1,...,T$}{
Update $f_t(\mathbf{x})$ using the $M$ labeled samples in $L$\;}
\caption{The FW-MT-iGS ALR algorithm.} \label{alg:fwMT-iGS}
\end{algorithm}

\subsection{FW-MT-GALR} \label{sect:fwMTGALR}

Although not mentioned in the original paper~\cite{Zhang2020}, GALR can also be used in both single-task ALR and multi-task ALR. Since GALR is almost identical to GSx, FW-MT-GALR is also almost identical to FW-MT-GSx, except that $d_{ij}^{\mathbf{x}^\mathrm{w}}$ in (\ref{eq:fwmtdnmx}) of FW-MT-GSx is replaced by:
\begin{align}
d_{ij}^{\mathbf{x}^\mathrm{w}}=\prod_{t=1}^T\left|\mathbf{x}_i^{\mathrm{w}_t}-\mathbf{x}_j^{\mathrm{w}_t}\right|,\quad \forall \mathbf{x}_i\in L, \forall \mathbf{x}_j\in U, \label{eq:fwmtdnmxGALR}
\end{align}
i.e., feature weighted $\mathcal{L}_1$ distance, instead of $\mathcal{L}_2$ distance, is used.

Due to the high similarity between FW-MT-GALR and FW-MT-GSx, and the page limit, the pseudo-code of FW-MT-GALR is omitted.

\subsection{Discussion}

There are some other popular active learning algorithms, e.g., active learning with Gaussian processes~\cite{Seo2000,Holzmueller2023}, Bayesian active learning by disagreement (BALD)~\cite{Houlsby2011}, etc., which use uncertainty (informativeness) as the primary criterion for sample selection, and may also be used for regression problems with little or no modifications. However, these approaches may not be enhanced by our proposed feature weighting strategy, as they do not involve computing the distances among different samples (which is typically used in the diversity and representativeness measures), so feature weighting is unemployable. One exception is the afore-introduced GALR, whose theoretical derivation starts from uncertainty reduction but the final formulation actually uses diversity as the sample selection criterion, so it can also be enhanced by feature weighting.

\section{Experiment Results on Single-Task ALR} \label{sect:experiment1}

Experiments are performed in this section to demonstrate the performance of the four proposed single-task feature weighted ALR approaches. The source code is available on \href{https://github.com/drwuHUST/FWALR}{github}.

\subsection{Datasets}

We used 11 single-task regression datasets from the UCI Machine Learning Repository\footnote{\url{http://archive.ics.uci.edu/ml/index.php}} and the CMU StatLib Datasets Archive\footnote{\url{http://lib.stat.cmu.edu/datasets/}}, which have been used in previous ALR studies~\cite{Yu2010,Cai2013,Cai2017,drwuSAL2019,drwuiGS2019}. Table~\ref{tab:datasets} summarizes their main characteristics. The autoMPG, CPS and BikeSharing datasets contain both numerical and categorical features. One-hot encoding was used to covert the categorical values into numerical ones. Particularly, for BikeSharing, we removed the Day, Month and Weekday features, as their one-hot codings are too long, and their information can be largely represented by other features.

For each dataset, each feature dimension was independently $z$-score normalized. No other preprocessing was applied.

\begin{table}[htpb]
\caption{Summary of the 11 single-task regression datasets.} \label{tab:datasets}
\centering \setlength{\tabcolsep}{.8mm}
\begin{threeparttable}
\begin{tabular}{l|cccccc}   \hline
 & &  \# of         &  \# of    &   \# of   &   \# of  & \# of           \\
Dataset & Source& samples        & raw   &   numerical  &   categorical  & total           \\
& &  &  features  &   features &   features &  features       \\ \hline
Yacht\tnote{1}  &  UCI &   308  &   6&6 & 0& 6\\
autoMPG\tnote{2}  & UCI &   392 & 7  &    6     & 1 & 9 \\
NO2\tnote{3}  &  StatLib &   500  &     7 & 7 & 0   &  7  \\
PM10\tnote{3}  &  StatLib &   500  &     7 & 7 & 0   &  7  \\
Housing\tnote{4} &  UCI & 506 &  13 & 13 & 0    &  13 \\
CPS\tnote{5}  &   StatLib & 534  &   11 & 8 &3     &  19 \\
EE-Cooling\tnote{6}    &UCI     &768    &8  &8  &0     &8\\
Concrete\tnote{7}& UCI & 1,030 & 8 & 8 & 0 & 8  \\
Airfoil\tnote{8} & UCI & 1,503 & 5 & 5 & 0 & 5 \\
Wine-white\tnote{9} & UCI & 4,898 &  11 & 11 & 0    &   11 \\
BikeSharing\tnote{10} & UCI & 17,389 &  13 & 7 & 6    &   39 \\  \hline
\end{tabular}
  \begin{tablenotes}
\item[1] \url{https://archive.ics.uci.edu/ml/datasets/Yacht+Hydrodynamics}
\item[2] \url{https://archive.ics.uci.edu/ml/datasets/auto+mpg}
\item[3] \url{http://lib.stat.cmu.edu/datasets/}
\item[4] \url{https://archive.ics.uci.edu/ml/machine-learning-databases/housing/}
\item[5] \url{http://lib.stat.cmu.edu/datasets/CPS_85_Wages}
\item[6] \url{http://archive.ics.uci.edu/ml/datasets/energy+efficiency}
\item[7] \url{https://archive.ics.uci.edu/ml/datasets/Concrete+Compressive+Strength}
\item[8] \url{https://archive.ics.uci.edu/ml/datasets/Airfoil+Self-Noise}
\item[9] \url{https://archive.ics.uci.edu/ml/datasets/Wine+Quality}
\item[10] \url{https://archive.ics.uci.edu/dataset/275/bike+sharing+datase}
  \end{tablenotes}
\end{threeparttable}
\end{table}

\subsection{Algorithms}

We compared the performance of the following 11 single-task ALR approaches on the 11 datasets in Table~\ref{tab:datasets}:
\begin{enumerate}
\item \texttt{BL}, which is a baseline approach that randomly selects all $M$ samples to label.
\item \texttt{EMCM} (expected model change maximization)~\cite{Cai2013}, which considers only the informativeness, represented by expected model parameter changes, in sample selection.
\item \texttt{QBC} (query-by-committee)~\cite{RayChaudhuri1995}, which considers only the informativeness, represented by model uncertainty, in sample selection.
\item \texttt{RD}, which is the RD ALR algorithm~\cite{drwuSAL2019} introduced in Section~\ref{sect:RD}.
\item \texttt{FW-RD}, which is our proposed FW-RD ALR algorithm introduced in Section~\ref{sect:fwRD}.
\item \texttt{GSx}, which is the GSx ALR algorithm~\cite{drwuiGS2019} introduced in Section~\ref{sect:GSx}.
\item \texttt{FW-GSx}, which is our proposed FW-GSx ALR algorithm introduced in Section~\ref{sect:fwGSx}.
\item \texttt{iGS}, which is the iGS ALR algorithm~\cite{drwuiGS2019} introduced in Section~\ref{sect:iGS}.
\item \texttt{FW-iGS}, which is our proposed FW-iGS ALR algorithm introduced in Section~\ref{sect:fwiGS}.
\item \texttt{GALR}, which is the GALR algorithm~\cite{Zhang2020} introduced in Section~\ref{sect:GALR}.
\item \texttt{FW-GALR}, which is our proposed FW-GALR algorithm introduced in Section~\ref{sect:fwGALR}.
\end{enumerate}
All 11 approaches constructed a ridge regression model from the labeled samples. $\lambda=0.1$ was used in the objective function $\underset{\boldsymbol{\beta}}{\operatorname{min}} (\| \mathbf{y}-\mathbf{X} \boldsymbol{\beta} \|^2 + \lambda \|\boldsymbol{\beta}\|^2)$, where $\boldsymbol{\beta}$ contains the ridge regression bias and coefficients.

\subsection{Performance Evaluation Process} \label{sect:process}

The evaluation process for single-task ALR approaches was identical to that used in~\cite{drwuSAL2019}. For each dataset, let $\mathbf{P}$ be the pool of all samples. We first randomly selected 80\% of the total samples as our training pool (denoted as $\mathbf{P}_{80}$), initialized the first $d+1$ labeled samples (for each dataset, $d$ was the number of total features in Table~\ref{tab:datasets}), identified one sample to label in each subsequent iteration by different algorithms, and built a ridge regression model. The maximum number of samples to be labeled, $M$, was 10\% of the size of $\mathbf{P}_{80}$. For large datasets, we clipped $M$ at 60.

After each iteration of each algorithm, we computed the root mean squared error (RMSE) and the correlation coefficient (CC) on the test set as the performance measures. As pointed out in~\cite{drwuSAL2019}, RMSE should be considered as the primary performance measure, because it is directly optimized in the objective function of ridge regression, whereas CC is not. Generally, as RMSE decreases, CC increases, but there could be exceptions.

\subsection{Main Results}

\begin{figure*}[htpb]
\subfigure[]{\label{fig:dataset1}     \includegraphics[width=.32\linewidth,clip]{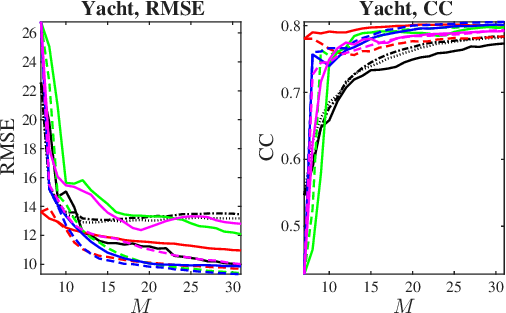}}
\subfigure[]{\label{fig:dataset2}     \includegraphics[width=.32\linewidth,clip]{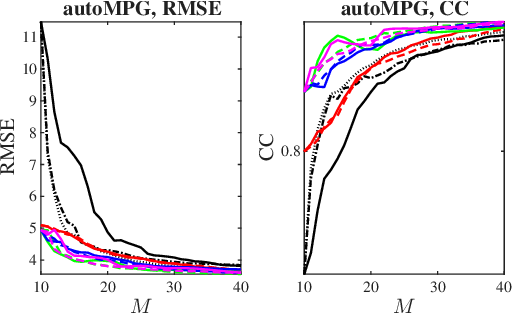}}
\subfigure[]{\label{fig:dataset3}     \includegraphics[width=.32\linewidth,clip]{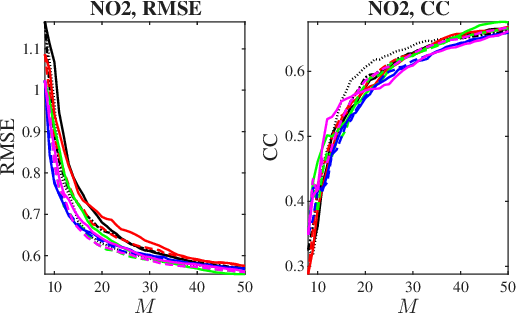}}
\subfigure[]{\label{fig:dataset4}     \includegraphics[width=.32\linewidth,clip]{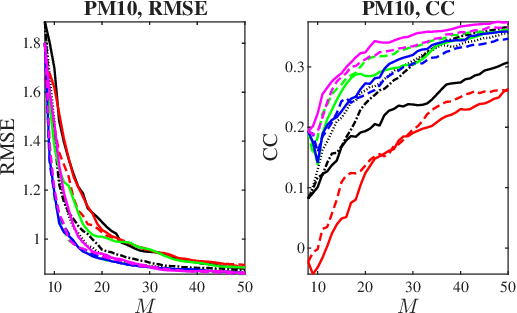}}
\subfigure[]{\label{fig:dataset5}     \includegraphics[width=.32\linewidth,clip]{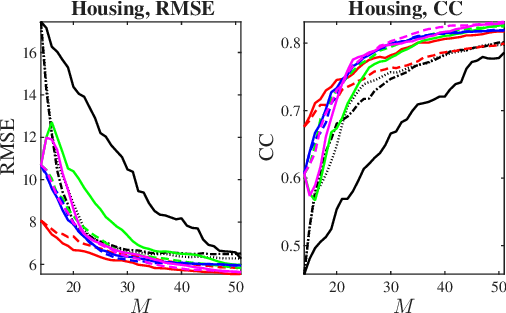}}
\subfigure[]{\label{fig:dataset6}     \includegraphics[width=.32\linewidth,clip]{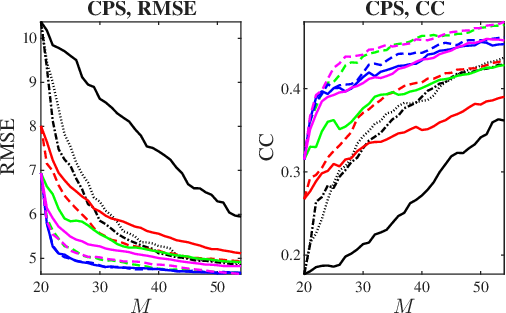}}
\subfigure[]{\label{fig:dataset7}     \includegraphics[width=.32\linewidth,clip]{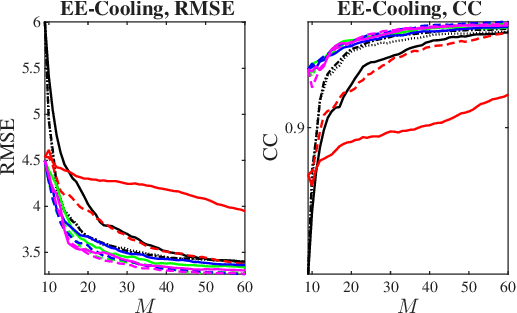}}
\subfigure[]{\label{fig:dataset8}     \includegraphics[width=.32\linewidth,clip]{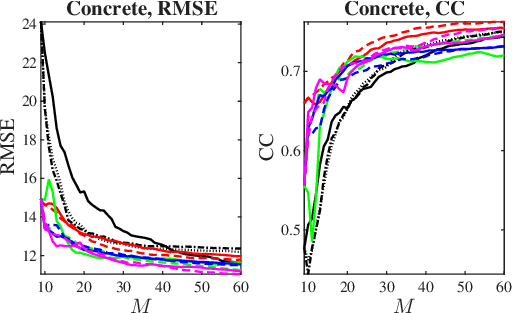}}
\subfigure[]{\label{fig:dataset9}     \includegraphics[width=.32\linewidth,clip]{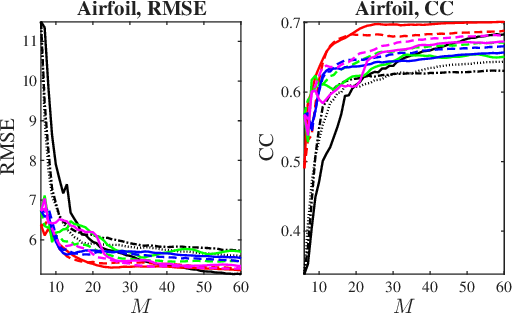}}
\subfigure[]{\label{fig:dataset10}     \includegraphics[width=.32\linewidth,clip]{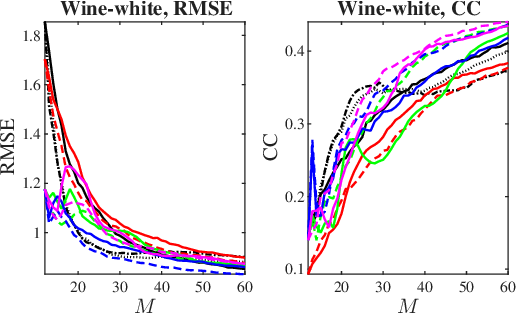}}
\subfigure[]{\label{fig:dataset11}     \includegraphics[width=.42\linewidth,clip]{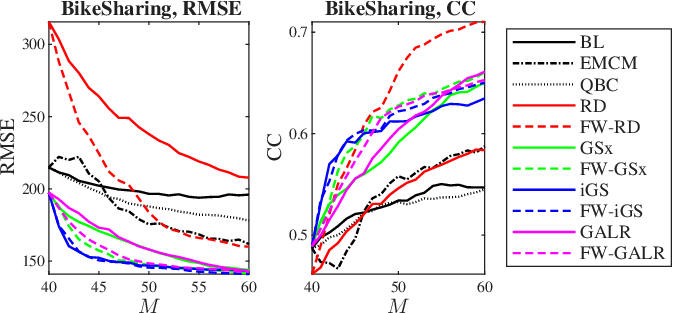}}
\caption{Performance of the 11 single-task ALR algorithms on the 11 datasets, averaged over 100 runs. (a) Yacht; (b) autoMPG; (c) NO2; (d) PM110; (e) Housing; (f) CPS; (g) EE-Cooling; (h) Concrete; (i) Airfoil; (j) Wine-white; and, (k) BikeSharing.} \label{fig:results12}
\end{figure*}

Fig.~\ref{fig:results12} shows the RMSEs and CCs of the 11 single-task ALR algorithms on the 11 datasets, averaged over 100 runs. We can observe that:
\begin{enumerate}
\item Generally, as $M$ (the number of labeled samples) increased, all 11 algorithms achieved better performance (smaller RMSE and larger CC), which is intuitive.
\item \texttt{EMCM}, \texttt{QBC}, \texttt{RD}, \texttt{GSx}, \texttt{iGS} and \texttt{GALR} outperformed \texttt{BL} on almost all datasets, suggesting the effectiveness of ALR.
\item Generally, the feature weighed ALR approaches outperformed their unweighted counterparts, i.e., \texttt{FW-RD} outperformed \texttt{RD}, \texttt{FW-GSx} outperformed \texttt{GSx}, \texttt{FW-iGS} outperformed \texttt{iGS}, and \texttt{FW-GALR} outperformed \texttt{GALR}, demonstrating the effectiveness of feature weighting.
\end{enumerate}

To see the forest for the trees, as in~\cite{drwuSAL2019}, we also define an aggregated performance measure called the area under the curve (AUC) for the average RMSE and the average CC on each of the 11 datasets in Fig.~\ref{fig:results12}. The AUC of \texttt{BL} was used to normalize the AUCs of the other 10 ALR algorithms for each dataset (so the AUC of \texttt{BL} was always 1). For the RMSE, a smaller AUC is preferred; for the CC, a larger AUC is preferred. Table~\ref{tab:AUCs} gives the AUC values. We can observe that:
\begin{enumerate}
\item On average, all 10 ALR approaches achieved smaller RMSEs and larger CCs than \texttt{BL}, suggesting the effectiveness of ALR.
\item On average, the feature weighed ALR approaches outperformed their unweighted counterparts, demonstrating again the effectiveness of feature weighting.
\item Generally, feature weighting maintained the rank of the original ALR approaches. For example, in the average normalized AUCs of the RMSEs of the unweighted ALR approaches were in the order of \texttt{RD}$\ge$\texttt{GSx}$>$\texttt{GALR}$>$\texttt{iGS} (from worst to best), and the average normalized AUCs of the RMSEs of the feature weighted ALR approaches were in the order of \texttt{FW-RD}$>$\texttt{FW-GSx}$\ge$\texttt{FW-GALR}$>$\texttt{FW-iGS} (from worst to best). The CCs showed a similar pattern.
\item Overall, in terms of RMSE, the primary performance matric, \texttt{iGS} was the best-performing single-task ALR approach without feature weighting, and the corresponding \texttt{FW-iGS} was the best-performing feature weighted single-task ALR approach.
\end{enumerate}

\begin{table*}[htpb]
\caption{Normalized AUCs of RMSEs and CCs of different algorithms on the 11 datasets.} \label{tab:AUCs}
\centering \setlength{\tabcolsep}{.8mm}
\begin{tabular}{l|ccccccccccc|ccccccccccc}   \hline
\multirow[c]{3}{*}{Dataset}&\multicolumn{11}{c}{RMSE} &\multicolumn{11}{|c}{CC}\\ \cline{2-23}
& & & &  & \texttt{FW} & & \texttt{FW} &  & \texttt{FW} & & \texttt{FW}  & && & & \texttt{FW} &  & \texttt{FW} & & \texttt{FW} &  & \texttt{FW}   \\
 & \texttt{BL} & \texttt{EMCM}& \texttt{QBC} & \texttt{RD} & \texttt{-RD} & \texttt{GSx} & \texttt{-GSx}  &\texttt{iGS} & \texttt{-iGS}  & \texttt{GALR} & \texttt{-GALR}& \texttt{BL} & \texttt{EMCM}& \texttt{QBC} & \texttt{RD} & \texttt{-RD} & \texttt{GSx} & \texttt{-GSx}  & \texttt{iGS} & \texttt{-iGS}  & \texttt{GALR} & \texttt{-GALR} \\ \hline
Yacht       &  1.00& 1.14& 1.12& 0.97& 0.88& 1.22& 0.97& 0.95&  0.91&  1.17& 0.99& 1.00& 1.02& 1.02&  1.10& 1.07& 1.04& 1.05& 1.06& 1.07& 1.05& 1.05\\
autoMPG     &  1.00& 0.87& 0.84& 0.79& 0.79& 0.73& 0.72& 0.76&  0.75&  0.75& 0.73& 1.00& 1.01& 1.02&  1.03& 1.02& 1.05& 1.05& 1.04& 1.04& 1.05& 1.05\\
NO2         &  1.00& 0.97& 0.95& 1.00& 0.98& 0.95& 0.93& 0.93&  0.93&  0.94& 0.93& 1.00& 1.00& 1.02&  1.00& 1.01& 1.01& 1.00& 0.99& 0.98& 1.00& 1.01\\
PM10        &  1.00& 0.93& 0.92& 0.99& 0.97& 0.96& 0.89& 0.89&  0.89&  0.92& 0.89& 1.00& 1.22& 1.26&  0.69& 0.78& 1.34& 1.40& 1.32& 1.27& 1.49& 1.43\\
Housing     &  1.00& 0.72& 0.72& 0.59& 0.62& 0.75& 0.67& 0.65&  0.65&  0.68& 0.64& 1.00& 1.10& 1.11&  1.18& 1.15& 1.15& 1.18& 1.18& 1.18& 1.17& 1.18\\
CPS         &  1.00& 0.74& 0.77& 0.75& 0.70& 0.68& 0.63& 0.62&  0.62&  0.66& 0.63& 1.00& 1.37& 1.38&  1.32& 1.47& 1.50& 1.70& 1.63& 1.65& 1.62& 1.71\\
EE-Cooling  &  1.00& 0.96& 0.96& 1.11& 0.97& 0.93& 0.91& 0.93&  0.91&  0.91& 0.90& 1.00& 1.01& 1.01&  0.97& 1.00& 1.01& 1.01& 1.01& 1.01& 1.01& 1.01\\
Concrete    &  1.00& 0.95& 0.95& 0.91& 0.89& 0.87& 0.86& 0.86&  0.87&  0.84& 0.84& 1.00& 1.00& 1.00&  1.06& 1.07& 1.02& 1.04& 1.03& 1.02& 1.04& 1.05\\
Airfoil     &  1.00& 1.03& 1.01& 0.90& 0.91& 0.98& 0.94& 0.95&  0.94&  0.95& 0.92& 1.00& 0.98& 0.99&  1.11& 1.09& 1.03& 1.06& 1.04& 1.05& 1.05& 1.07\\
Wine-white  &  1.00& 0.95& 0.94& 1.04& 1.00& 0.94& 0.92& 0.91&  0.87&  0.96& 0.93& 1.00& 1.00& 1.02&  0.89& 0.86& 0.92& 1.04& 1.00& 1.08& 1.04& 1.10\\
BikeSharing &  1.00& 0.85& 0.92& 1.11& 0.90& 0.76& 0.74& 0.74&  0.73&  0.75& 0.74& 1.00& 1.07& 0.99&  1.05& 1.24& 1.17& 1.18& 1.15& 1.17& 1.18& 1.18\\ \hline
Average     &  1.00& 0.92& 0.92& 0.92& 0.87& 0.89& 0.83& 0.84&  \textbf{0.82}&  0.87& 0.83& 1.00& 1.07& 1.07&  1.04& 1.07& 1.11& 1.16& 1.13& 1.14& 1.15& \textbf{1.17}\\
\hline
\end{tabular}
\end{table*}

To test if the performance improvements brought by feature weighting were statistically significant, Wilcoxon paired signed-rank tests \cite{Wilcoxon1945} were conducted on the AUCs across the 11 datasets. The $p$-values are shown in Table~\ref{tab:pAUCs}. For the primary performance measure, RMSE, the performance differences between the feature weighted ALRs and their unweighted counterparts were always statistically significant ($p<0.05$); moreover, all four feature weighted ALRs statistically significantly outperformed the five basic approaches (\texttt{BL}, \texttt{EMCM}, \texttt{QBC}, \texttt{RD}, and \texttt{GSx}).

\begin{table*}[htpb]
\caption{$p$-values of Wilcoxon paired signed-rank test on the AUCs of RMSEs and CCs across the 11 datasets. $p$-values smaller than 0.05 are in bold.} \label{tab:pAUCs}
\centering \setlength{\tabcolsep}{.8mm}
\begin{tabular}{l|cccccccccc|cccccccccc}   \hline
\multirow[c]{2}{*}{Approach}&\multicolumn{10}{c}{RMSE} &\multicolumn{10}{|c}{CC}\\ \cline{2-21}
 & \texttt{BL} & \texttt{EMCM}& \texttt{QBC} & \texttt{RD} & \texttt{FW-RD} & \texttt{GSx} & \texttt{FW-GSx} & \texttt{iGS} & \texttt{FW-iGS}
 &\texttt{GALR}   & \texttt{BL} & \texttt{EMCM}& \texttt{QBC} & \texttt{RD} & \texttt{FW-RD} & \texttt{GSx} & \texttt{FW-GSx} & \texttt{iGS} & \texttt{FW-iGS}
 & \texttt{GALR}    \\ \hline
\texttt{EMCM}    & \textbf{0.04}&    &     &      &     &     &     &     &     &     &  \textbf{0.01}&     &    &     &     &     &     &     &     &\\
\texttt{QBC}     & \textbf{0.03}&0.41&     &      &     &     &     &     &     &     &  \textbf{0.02}& 0.12&    &     &     &     &     &     &     &      \\
\texttt{RD}      & 0.28&1.00& 0.97&      &     &     &     &     &     &     &  0.37& 1.00&0.97&     &     &     &     &     &     &      \\
\texttt{FW-RD}   & \textbf{0.00}&0.21& 0.12&  \textbf{0.03}&     &     &     &     &     &     &  0.17& 0.41&0.58& 0.58&     &     &     &     &     &      \\
\texttt{GSx}     & \textbf{0.02}&0.17& 0.32&  0.46& 0.70&     &     &     &     &     &  \textbf{0.02}& \textbf{0.02}&0.05& 0.46& 0.97&     &     &     &     &      \\
\texttt{FW-GSx}  & \textbf{0.00}&\textbf{0.00}& \textbf{0.00}&  \textbf{0.03}& 0.10&  \textbf{0.00} &     &     &     &     &  \textbf{0.00}& \textbf{0.00}&\textbf{0.00}& 0.17& 0.52& \textbf{0.00}    &     &     &     &      \\
\texttt{iGS}    & \textbf{0.00}&\textbf{0.00}& \textbf{0.00}&  \textbf{0.03}& 0.10& \textbf{0.01}& 0.90&     &     &     &  \textbf{0.00}& \textbf{0.01}&\textbf{0.02}& 0.24& 0.58& 0.24& \textbf{0.02}&     &     &      \\
\texttt{FW-iGS} & \textbf{0.00}&\textbf{0.00}& \textbf{0.00}&  \textbf{0.02}& \textbf{0.03}& \textbf{0.00}& 0.58& \textbf{0.02}&     &     &  \textbf{0.00}& \textbf{0.00}&\textbf{0.01}& 0.24& 0.52& 0.21& 0.21& 0.32&     &      \\
\texttt{GALR}     & \textbf{0.02}&\textbf{0.01}& \textbf{0.03}&  0.24& 0.32& \textbf{0.03}& \textbf{0.01}& 0.10& \textbf{0.01}&     &  \textbf{0.00}& \textbf{0.00}&\textbf{0.00}& 0.21& 0.52& \textbf{0.01}& 0.32& 0.21& 0.97&      \\
\texttt{FW-GALR}  & \textbf{0.00}&\textbf{0.00}& \textbf{0.00}&  \textbf{0.01}& 0.08& \textbf{0.00}& 1.00& 0.58& 1.00& \textbf{0.00}&  \textbf{0.00}& \textbf{0.00}&\textbf{0.00}& 0.12& 0.28& \textbf{0.00}& \textbf{0.03}& \textbf{0.01}& \textbf{0.01}& 0.15\\ \hline
\end{tabular}
\end{table*}

\subsection{Computational Cost}

This subsection compares the computational cost of different ALR approaches, an important consideration in real-world applications.

\begin{figure}[htpb] \centering
\subfigure[]{\label{fig:compCostN}     \includegraphics[width=.48\linewidth,clip]{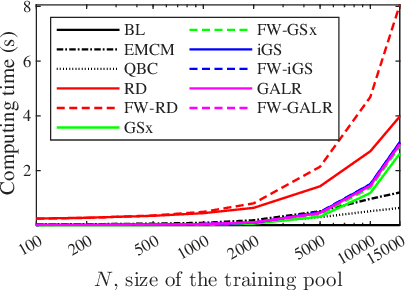}}
\subfigure[]{\label{fig:compCostD}     \includegraphics[width=.48\linewidth,clip]{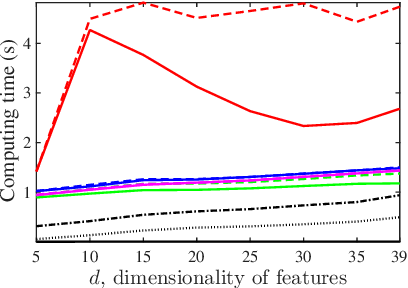}}
\caption{Computational cost of the 11 algorithms on BikeSharing, for $M\in[20,60]$, averaged over 100 runs. (a) $d=39$, while $N$ increases from 100 to 15,000; and, (b) $N=10,000$, while $d$ increases from 5 to 39.} \label{fig:compCost}
\end{figure}

Using the BikeSharing dataset as an example, Fig.~\ref{fig:compCostN} shows the total computing time (in seconds) when $M$ (the number of selected samples) increased from 20 to 60, for $d=39$ and different $N$ (total number of unlabelled samples). Fig.~\ref{fig:compCostD} shows the total computing time when $M$ increased from 20 to 60, for $N=10,000$ and different $d$ (feature dimensionality). Both were averaged over 100 runs. We can observe that:
\begin{enumerate}
\item The computational cost of the five existing ALR approaches are in the order of \texttt{QBC} $<$ \texttt{EMCM} $<$ \texttt{GSx} $<$ \texttt{GALR} $<$ \texttt{iGS} $<$ \texttt{RD}. \texttt{RD} has the highest computational cost, because it uses time-consuming $k$-means clustering in every iteration.
\item For the feature weighted ALR approaches, the computational cost is in the order of \texttt{FW-GSx} $<$ \texttt{FW-GALR} $<$ \texttt{FW-iGS} $<$ \texttt{FW-RD}, consistent with their unweighted counterparts.
\item Except for \texttt{FW-RD}, the extra computational burden introduced by feature weighting is very minor, sometimes almost unnoticeable (e.g., \texttt{FW-iGS} and \texttt{FW-GALR}).
\end{enumerate}

\subsection{Accuracy of the Estimated Feature Weights}

The feature weights are estimated from a very small number of labelled samples; it is interesting to know how accurate they are.

\begin{figure}[htpb]\centering
\includegraphics[width=.95\linewidth,clip]{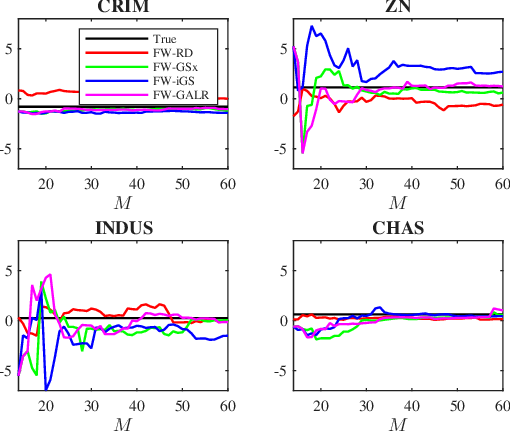}
\caption{True and estimated feature weights on Housing. Results for first four of the 13 features are shown.} \label{fig:RRcoef}
\end{figure}

Using the Housing dataset as an example, Fig.~\ref{fig:RRcoef} shows the true feature weights computed based on the entire training data pool (consisting of 80\% of the 506 samples), and the ridge regression estimated feature weights based on $M\in[14,60]$ labelled samples selected by different feature weighted ALR approaches. The feature weights estimated from all four feature weighted ALR approaches generally converged after a small $M$ (usually smaller than 30), and their magnitude approached the true weights, resulting good RMSE and CC performance, as indicated in Fig.~\ref{fig:dataset5} and Table~\ref{tab:AUCs}.

\subsection{Robustness to Correlated/Redundant Features}

This subsection studies how correlated/redundant features may impact the weight calculation and subsequently the ALR sample selection effectiveness.

Due to page limit, we only considered the first two datasets. For each dataset, we extracted its first two features and augmented them to the original feature set. For example, the original Yacht dataset has six features, and we augmented its first two features to the six original features, resulting in a total of eight features. In this way, significant feature correlation/redundancy was introduced. All ALR algorithms in the previous subsection were evaluated on the feature augmented datasets.

\begin{figure}[htpb] \centering
\subfigure[]{\label{fig:dataset1-redun}     \includegraphics[width=.34\linewidth,clip]{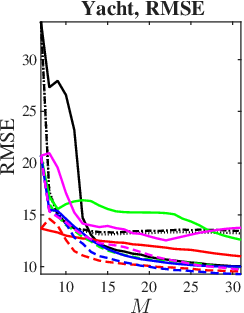}}
\subfigure[]{\label{fig:dataset2-redun}     \includegraphics[width=.58\linewidth,clip]{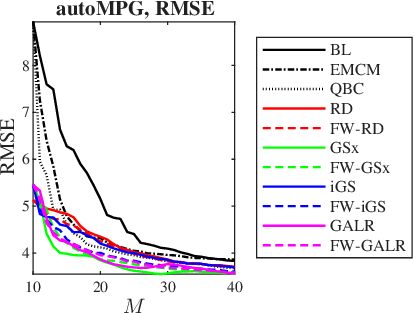}}
\caption{Performance of the 11 single-task ALR algorithms, averaged over 100 runs. (a) Yacht; and, (b) autoMPG. Two redundant features were added to each dataset.} \label{fig:resultsRedun}
\end{figure}

The RMSEs are shown in Fig.~\ref{fig:resultsRedun}. The normalized AUCs of RMSEs and CCs are shown in Table~\ref{tab:redundant}. After adding correlated/redundant features, the RMSEs of \texttt{BL} and the six unweighted ALR approaches (\texttt{EMCM}, \texttt{QBC}, \texttt{RD}, \texttt{GSx}, \texttt{iGS} and \texttt{GALR}) generally increased (more obvious on Yacht), indicating worse performance; however, the feature weighted ALRs (\texttt{FW-RD}, \texttt{FW-GSx}, \texttt{FW-iGS} and \texttt{FW-GALR}) still outperformed, or at least achieved comparable performance with, their unweighted counterparts. These results validated the effectiveness and robustness of feature weighted ALRs to feature correlation/redundancy.

\begin{table}[htpb]
\caption{Normalized AUCs of RMSE and CC before and after adding correlated/redundant features.} \label{tab:redundant}
\centering \setlength{\tabcolsep}{1mm}
\begin{tabular}{l|cc|cc|cc|cc}   \hline
& \multicolumn{4}{c|}{RMSE} & \multicolumn{4}{c}{CC} \\ \cline{2-9}
Algorithm & \multicolumn{2}{c|}{Yacht} & \multicolumn{2}{c|}{AutoMPG} & \multicolumn{2}{c|}{Yacht} & \multicolumn{2}{c}{AutoMPG} \\ \cline{2-9}
& Before & After & Before & After& Before & After& Before & After\\ \hline
\texttt{BL}      &  1.00&    1.18&    1.00&    0.95&    1.00&    0.98&    1.00&    1.00\\
\texttt{EMCM}    &  1.14&    1.19&    0.87&    0.85&    1.02&    1.02&    1.01&    1.01\\
\texttt{QBC}     &  1.12&    1.18&    0.84&    0.81&    1.02&    1.01&    1.02&    1.02\\
\texttt{RD}      &  0.97&    1.00&    0.79&    0.79&    1.10&    1.11&    1.03&    1.02\\
\texttt{FW-RD}   &  0.88&    0.88&    0.79&    0.78&    1.07&    1.08&    1.02&    1.02\\
\texttt{GSx}     &  1.22&    1.25&    0.73&    0.72&    1.04&    1.07&    1.05&    1.05\\
\texttt{FW-GSx}  &  0.97&    0.98&    0.72&    0.74&    1.05&    1.06&    1.05&    1.04\\
\texttt{iGS}     &  0.95&    0.97&    0.76&    0.78&    1.06&    1.07&    1.04&    1.03\\
\texttt{FW-iGS}  &  0.91&    0.92&    0.75&    0.75&    1.07&    1.08&    1.04&    1.04\\
\texttt{GALR}    &  1.17&    1.19&    0.75&    0.74&    1.05&    1.02&    1.05&    1.04\\
\texttt{FW-GALR} &  0.99&    0.98&    0.73&    0.75&    1.05&    1.06&    1.05&    1.03\\ \hline
\end{tabular}
\end{table}

However, these results do not exclude the possibility that feature selection~\cite{Cacciarelli2023} or transformation may be used to reduce the feature correlation/redundancy, and hence to improve the performance and efficiency of all ALR algorithms.

\subsection{Robustness to Irrelevant Features}

This subsection studies how irrelevant features may impact the weight calculation and subsequently the ALR sample selection effectiveness.

Due to page limit, we only considered the first two datasets. For each dataset, we added two irrelevant features to the original feature set: one is the index of the sample (e.g., from 1 to 308 for Yacht), and the other is random Gaussian noise. All ALR algorithms in the previous subsection were evaluated on the feature augmented datasets.

\begin{figure}[htpb]
\subfigure[]{\includegraphics[width=.34\linewidth,clip]{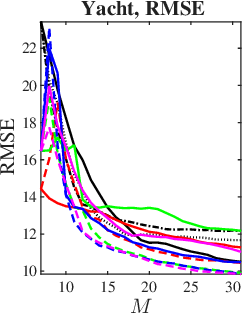}}
\subfigure[]{\includegraphics[width=.58\linewidth,clip]{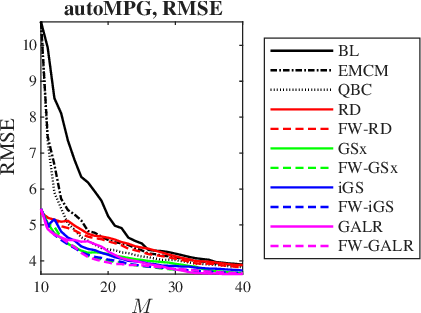}}
\caption{Performance of the 11 single-task ALR algorithms, averaged over 100 runs. (a) Yacht; and, (b) autoMPG. Two irrelevant features were added to each dataset.} \label{fig:resultsIrrel}
\end{figure}

The results are shown in Fig.~\ref{fig:resultsIrrel}. The normalized AUCs of RMSEs and CCs are shown in Table~\ref{tab:irrelevant}. Irrelevant feature worsen the performance of \texttt{BL}, but did not noticeably affect the effectiveness of \texttt{FW-RD}, \texttt{FW-GSx}, \texttt{FW-iGS} and \texttt{FW-GALR}: they still outperformed, or at least achieved comparable performance with, their unweighted counterparts. These results validated the effectiveness and robustness of feature weighted ALRs to irrelevant features.

\begin{table}[htpb]
\caption{Normalized AUCs of RMSE and CC before and after adding irrelevant features.} \label{tab:irrelevant}
\centering \setlength{\tabcolsep}{1mm}
\begin{tabular}{l|cc|cc|cc|cc}   \hline
& \multicolumn{4}{c|}{RMSE} & \multicolumn{4}{c}{CC} \\ \cline{2-9}
Algorithm & \multicolumn{2}{c|}{Yacht} & \multicolumn{2}{c|}{AutoMPG} & \multicolumn{2}{c|}{Yacht} & \multicolumn{2}{c}{AutoMPG} \\ \cline{2-9}
& Before & After & Before & After& Before & After& Before & After\\ \hline
\texttt{BL}      &      1.00&    1.12&    1.00&    1.01&    1.00&    0.93&    1.00&    0.98\\
\texttt{EMCM}    &      1.14&    1.12&    0.87&    0.90&    1.02&    0.96&    1.01&    1.01\\
\texttt{QBC}     &      1.12&    1.11&    0.84&    0.86&    1.02&    0.96&    1.02&    1.01\\
\texttt{RD}      &      0.97&    1.02&    0.79&    0.84&    1.10&    1.05&    1.03&    1.00\\
\texttt{FW-RD}   &      0.88&    1.01&    0.79&    0.82&    1.07&    0.94&    1.02&    1.00\\
\texttt{GSx}     &      1.22&    1.13&    0.73&    0.78&    1.04&    1.00&    1.05&    1.04\\
\texttt{FW-GSx}  &      0.97&    0.97&    0.72&    0.76&    1.05&    1.04&    1.05&    1.03\\
\texttt{iGS}     &      0.95&    1.03&    0.76&    0.78&    1.06&    1.03&    1.04&    1.02\\
\texttt{FW-iGS}  &      0.91&    0.97&    0.75&    0.76&    1.07&    1.04&    1.04&    1.03\\
\texttt{GALR}    &      1.17&    1.08&    0.75&    0.77&    1.05&    1.01&    1.05&    1.04\\
\texttt{FW-GALR} &      0.99&    0.95&    0.73&    0.76&    1.05&    1.03&    1.05&    1.04\\  \hline
\end{tabular}
\end{table}

%
%
%

\subsection{Effect of Different Number of Initially Labeled Samples}

Let $n_{\min}$ be the number of initially labeled samples. Though in all previous single-task ALR experiments, we always selected the first $n_{\min}=d+1$ unlabeled samples using RD or GSx, and then labeled them to initialize a ridge regression model, it should be noted that $n_{\min}$ can start from fewer than $d+1$ samples, and this does not impact the effectiveness of feature weighting.

\begin{figure}[htpb]\centering
\subfigure[]{\label{fig:minNs1} \includegraphics[width=.43\linewidth,clip]{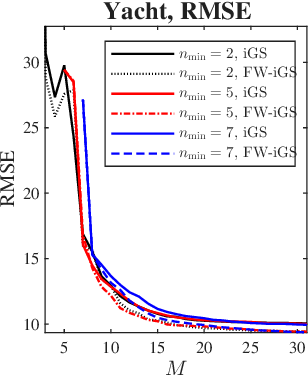}}
\subfigure[]{\label{fig:minNs2} \includegraphics[width=.45\linewidth,clip]{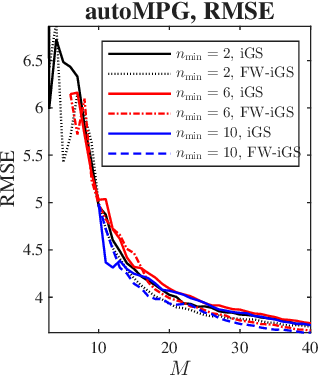}}
\caption{Effect of different number of initially labeled samples, $n_{\min}$ (averaged over 100 runs).  (a) Yacht, where $d=6$; and, (b) autoMPG, where $d=9$. } \label{fig:minNs}
\end{figure}

Fig.~\ref{fig:minNs} shows the RMSEs and CCs of \texttt{iGS} and \texttt{FW-iGS} (other algorithms were omitted, so that the curves are distinguishable) on two representative datasets when $n_{\min}$ increased from 2 to $d+1$. We can observe that:
\begin{enumerate}
\item \texttt{FW-iGS} always outperformed the corresponding \texttt{iGS}, regardless of $n_{\min}$, suggesting again the effectiveness and robustness of feature weighting.
\item There was no consistent pattern on whether a smaller or larger $n_{\min}$ was better. For example, Fig.~\ref{fig:minNs1} shows that a smaller $n_{\min}$ may result in a smaller RMSE for \texttt{FW-iGS}, whereas Fig.~\ref{fig:minNs2} shows that a smaller $n_{\min}$ may result in the largest RMSE for \texttt{FW-iGS}.
\item When $M<d+1$, both RMSE and CC exhibited large fluctuations as $M$ changed. This is because ridge regression (and consequently the feature weights) is unstable when there are too few labeled samples (this is also the reason why we started from $n_{\min}=d+1$ in previous experiments). A remedy for small $n_{\min}$ is to choose a larger regularization parameter $\lambda$, as done in the next subsection.
\end{enumerate}

In summary, our proposed feature weighted \texttt{FW-RD}, \texttt{FW-GSx}, \texttt{FW-iGS} and \texttt{FW-GALR}, and their original versions, can work with any $n_{\min}\ge 2$, though the optimal $n_{\min}$ is application dependent.

\subsection{Effect of Different Initialization Approaches}

More sophisticated initial sample selection approaches, such as iterative representativeness-diversity maximization (iRDM)~\cite{drwuiRDM2021} and informativeness-representativeness-diversity (IRD)~\cite{drwuIRD2021}, could also be used to improve the ALR performance, and they are independent of the proposed feature weighting scheme.

\begin{figure}[htpb]\centering
\includegraphics[width=.95\linewidth,clip]{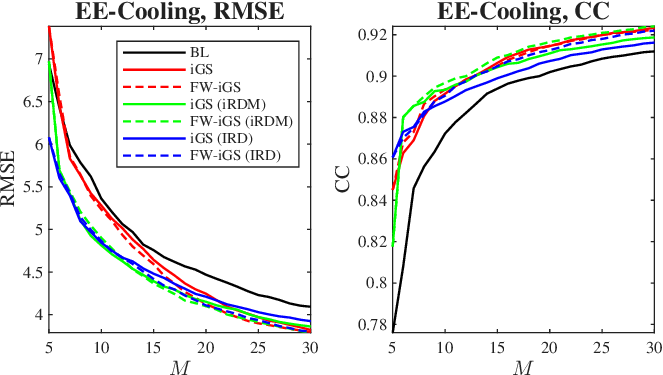}
\caption{Performance of different initialization approaches on the EE-Cooling dataset, averaged over 100 runs. \texttt{iGS (iRDM)} and \texttt{FW-iGS (iRDM)} used iRDM~\cite{drwuiRDM2021} to select the first 5 samples to label. \texttt{iGS (IRD)} and \texttt{FW-iGS (IRD)} used IRD~\cite{drwuIRD2021} to select the first 5 samples to label.} \label{fig:init}
\end{figure}

Fig.~\ref{fig:init} shows an example on the EE-cooling dataset, which had 7 features (i.e., $d=7$), but we started from 5 initial samples selected by GSx, iRDM and IRD, respectively. Since the number of initially labeled training samples was smaller than the feature dimensionality, we used a large $\lambda=10$ for strong regularization.

The results show that different initialization approaches resulted in different regression performance at the beginning: IRD was the best, and iRDM was the second best, both outperforming GSx. However, as $M$ increased, their performance gradually approached each other. More importantly, feature weighting was always effective, regardless of which initialization approach was used, and how small the initial $M$ was.

These results suggest that initial sample selection and feature weighting in subsequent sample selection are decoupled and complementary to each other, so a better initial sample selection approach can be directly plugged in to improve the overall ALR performance.

\subsection{Effectiveness on Different Regression Models}

In all previous experiments, the simple linear ridge regression model was used in both feature weight estimation and unlabeled sample prediction. This subsection studies two important questions:
\begin{enumerate}
\item Whether the feature weights estimated by a linear model is effective in nonlinear predictors?
\item Whether the feature weights estimated by a nonlinear model is more effective than those by a linear model?
\end{enumerate}

\begin{figure*}[htpb]
\subfigure[]{\label{fig:dataset1-Tree}     \includegraphics[width=.29\linewidth,clip]{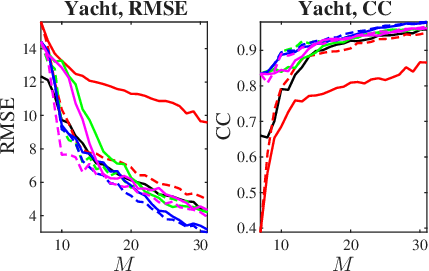}}
\subfigure[]{\label{fig:dataset2-Tree}     \includegraphics[width=.29\linewidth,clip]{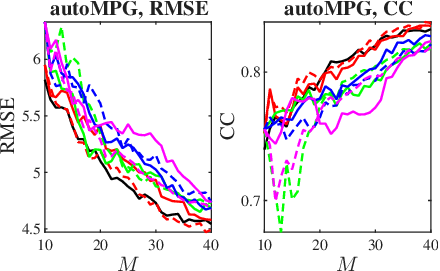}}
\subfigure[]{\label{fig:dataset3-Tree}     \includegraphics[width=.4\linewidth,clip]{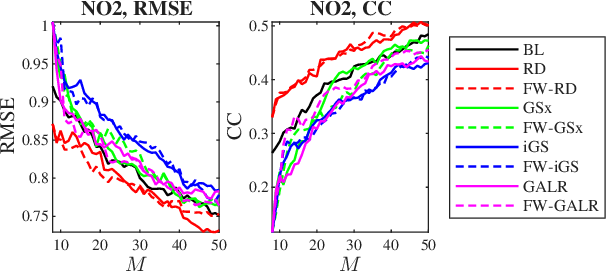}}
\caption{Performance of the nine single-task ALR algorithms, averaged over 200 runs. Linear ridge regression was used to estimate the feature weights in \texttt{FW-RD}, \texttt{FW-GSx}, \texttt{FW-iGS} and \texttt{FW-GALR}, and a nonlinear regression tree was trained for new sample prediction. (a) Yacht; (b) autoMPG; and, (c) NO2.} \label{fig:resultsTree}
\end{figure*}

Fig.~\ref{fig:resultsTree} shows the results when linear ridge regression was used to estimate the feature weights in \texttt{FW-RD}, \texttt{FW-GSx}, \texttt{FS-iGS} and \texttt{FW-GALR} and a nonlinear regression tree was trained for new sample label prediction. The feature weighted ALR algorithms generally achieved better performance than or comparable performance with their unweighted counterparts, i.e., the feature weights estimated by a linear model is also effective in nonlinear predictors.

\begin{figure*}[htpb]
\subfigure[]{\label{fig:dataset1-Tree2}     \includegraphics[width=.29\linewidth,clip]{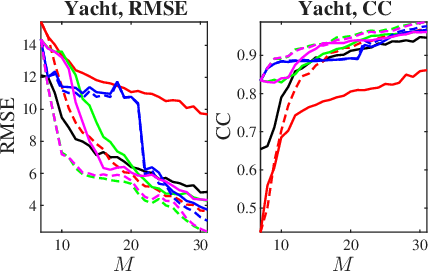}}
\subfigure[]{\label{fig:dataset2-Tree2}     \includegraphics[width=.29\linewidth,clip]{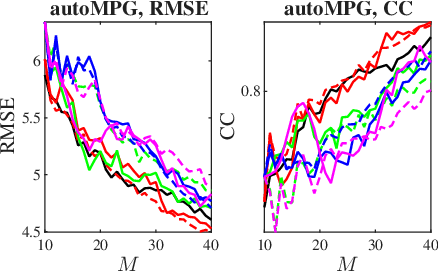}}
\subfigure[]{\label{fig:dataset3-Tree2}     \includegraphics[width=.4\linewidth,clip]{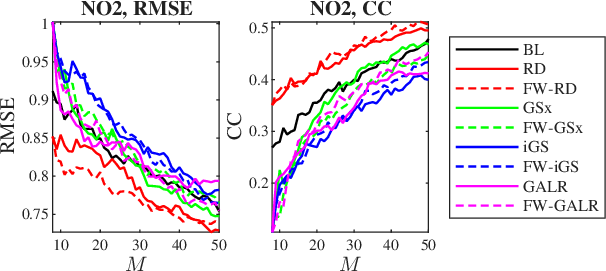}}
\caption{Performance of the nine single-task ALR algorithms, averaged over 200 runs. A nonlinear regression tree was used in both feature weight estimation in \texttt{FW-RD}, \texttt{FW-GSx}, \texttt{FW-iGS} and \texttt{FW-GALR}, and for new sample prediction. (a) Yacht; (b) autoMPG; and, (c) NO2.} \label{fig:resultsTree2}
\end{figure*}

Fig.~\ref{fig:resultsTree2} shows the results when a nonlinear regression tree was used for both estimating the feature weights in \texttt{FW-RD}, \texttt{FW-GSx}, \texttt{FS-iGS} and \texttt{FW-GALR}, and new sample label prediction. Similar to Fig.~\ref{fig:resultsTree}, the feature weighted ALR algorithms generally achieved better performance than or comparable performance with their unweighted counterparts. Comparing Figs.~\ref{fig:resultsTree} and \ref{fig:resultsTree2}, we cannot observe significant performance differences, i.e., the feature weights estimated by a linear model and a nonlinear model may be equally effective.

These experiments demonstrated the effectiveness and robustness of feature weighting: both linear and nonlinear regression models could be used in feature weight estimation, and for new sample label prediction. Moreover, the regression models used in these two parts could be different. In any case, feature weighting rarely hurts the ALR performance.

\section{Experiment Results on Multi-Task ALR} \label{sect:experiment2}

Experiments are performed in this section to demonstrate the performance of the three proposed feature weighted multi-task ALR approaches. The source code is available on \href{https://github.com/drwuHUST/FWALR}{github}.

\subsection{Datasets}

We used two multi-task regression datasets:
\begin{enumerate}
\item \emph{VAM}, which contains spontaneous speech with authentic emotions recorded from guests in a German TV talk-show and  has been used in many studies~\cite{Grimm2007a,Grimm2007b,Grimm2008,drwuICME2010,drwuInterSpeech2010,drwuMTALR2022}. There are 947 emotional utterances from 47 speakers (11m/36f), each labeled in the 3D space of valence, arousal and dominance. The same $46$ acoustic features extracted in previous research~\cite{drwuICME2010,drwuInterSpeech2010,drwuMTALR2022}, including nine pitch features, five duration features, six energy features, and 26 Mel Frequency Cepstral Coefficient (MFCC) features, were used in this paper. Each feature was normalized to mean 0 and standard deviation 1.
\item \emph{Energy Efficiency (EE)}\footnote{http://archive.ics.uci.edu/ml/datasets/energy+efficiency}, which aims to predict two outputs (the heating load and cooling load requirements) for 768 different building shapes simulated in Ecotect. There are eight numerical features, including relative compactness, surface area, wall area, roof area, overall height, orientation, glazing area, and glazing area distribution. EE-Cooling in Table~\ref{tab:datasets} is part of this dataset.
\end{enumerate}

As in the previous section, for each dataset, each feature dimension was independently $z$-score normalized. No other preprocessing was applied.

\subsection{Algorithms}

We compared the performance of the following seven multi-task ALR approaches on the VAM and EE datasets:
\begin{enumerate}
\item \texttt{BL}, which is a baseline approach that randomly selects all $M$ samples to label.
\item \texttt{GSx}, which is the GSx algorithm~\cite{drwuiGS2019} introduced in Section~\ref{sect:GSx}. It can be used in both single-task ALR and multi-task ALR.
\item \texttt{FW-MT-GSx}, which is our proposed FW-MT-GSx algorithm introduced in Section~\ref{sect:fwMTGSx}.
\item \texttt{MT-iGS}, which is the MT-iGS algorithm~\cite{drwuMTALR2022} introduced in Section~\ref{sect:MT-iGS}.
\item \texttt{FW-MT-iGS}, which is our proposed FW-MT-iGS algorithm introduced in Section~\ref{sect:fwMT-iGS}.
\item \texttt{GALR}, which is the GALR algorithm~\cite{Zhang2020} introduced in Section~\ref{sect:GALR}. It can be used in both single-task ALR and multi-task ALR.
\item \texttt{FW-MT-GALR}, which is our proposed FW-MT-GALR algorithm introduced in Section~\ref{sect:fwMTGALR}.
\end{enumerate}
All seven approaches construct a ridge regression model from the labeled samples, with ridge parameter $\lambda=10$. The ridge parameter here was much larger, as all seven approaches started from only 5 labeled samples, lower than the feature dimensionality, so a large $\lambda$ was needed to enforce strong regularization.

\subsection{Performance Evaluation Process}

The evaluation process for multi-task ALR approaches was identical to that used in~\cite{drwuMTALR2022}. For each dataset, 30\% of the samples were randomly selected as the training pool, and the remaining 70\% as the test pool. The number of selected samples ranged from 5 to 100 on VAM, and 5 to 40 on EE (the performance converged around $M=40$ on EE).

\subsection{Main Results}

\begin{figure}[htpb]\centering
\includegraphics[width=.95\linewidth,clip]{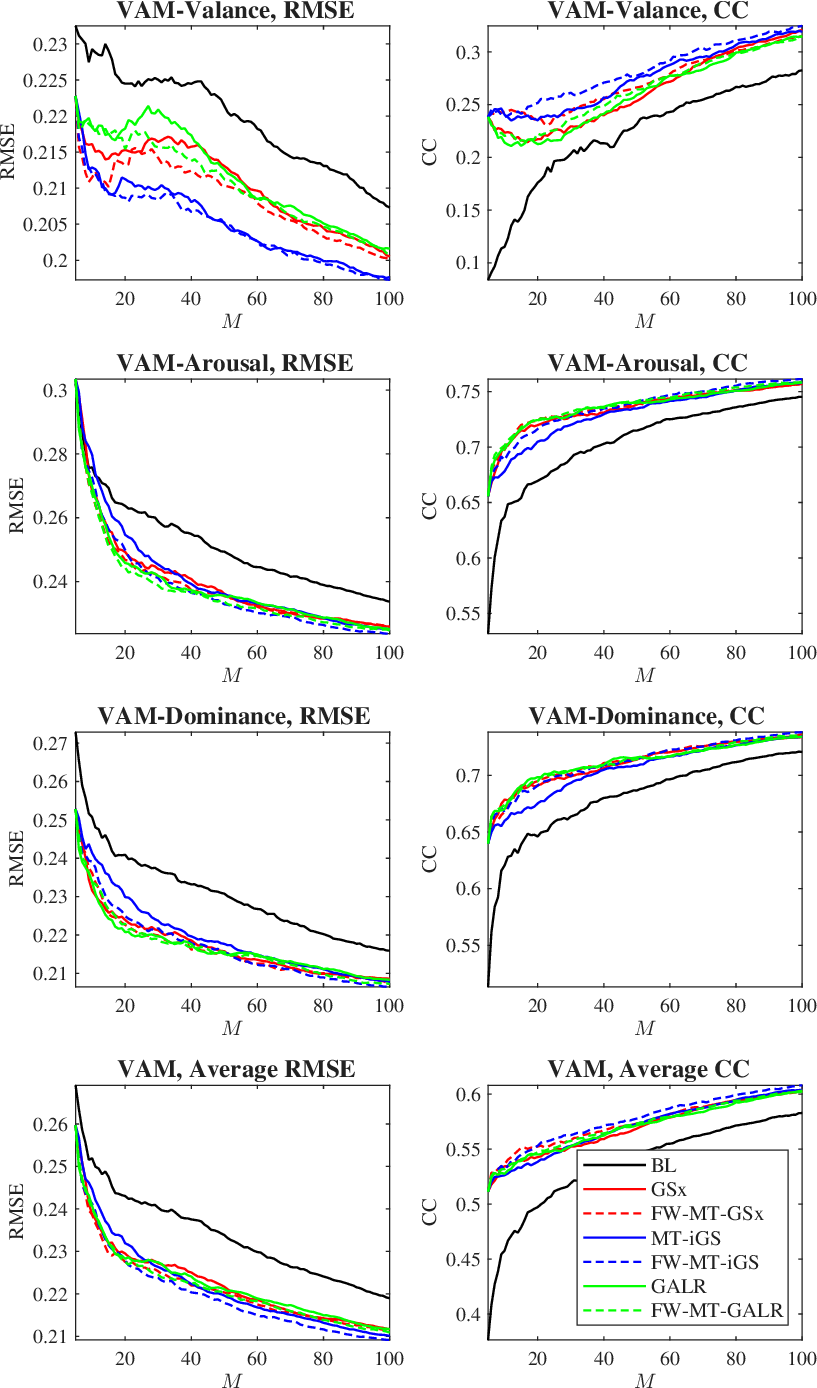}
\caption{Performance of the seven multi-task ALR algorithms on the VAM dataset, averaged over 100 runs.} \label{fig:VAM}
\end{figure}

Fig.~\ref{fig:VAM} shows the RMSEs and CCs of the seven multi-task ALR algorithms on the VAM dataset, averaged over 100 runs. We can observe that:
\begin{enumerate}
\item Generally, as $M$ increased, all seven algorithms achieved better performance (smaller RMSE and larger CC), which is intuitive.
\item \texttt{GSx}, \texttt{MT-iGS} and \texttt{GALR} outperformed \texttt{BL} in each individual task, suggesting that all these ALR approaches were effective.
\item Generally, the feature weighed ALR approaches outperformed their unweighted counterparts, i.e., \texttt{FW-MT-GSx} outperformed \texttt{GSx}, \texttt{FW-MT-iGS} outperformed \texttt{MT-iGS}, and \texttt{FW-MT-GALR} outperformed \texttt{GALR}, demonstrating again the effectiveness of feature weighting.
\item Overall, \texttt{MT-iGS} was the best performing multi-task ALR approach without feature weighting, as it more comprehensively considers the diversities of the selected samples than \texttt{GSx} and \texttt{GALR}. The corresponding \texttt{FW-MT-iGS} was the best performing feature weighted multi-task ALR approach.
\end{enumerate}

\begin{figure}[htpb]\centering
\includegraphics[width=.95\linewidth,clip]{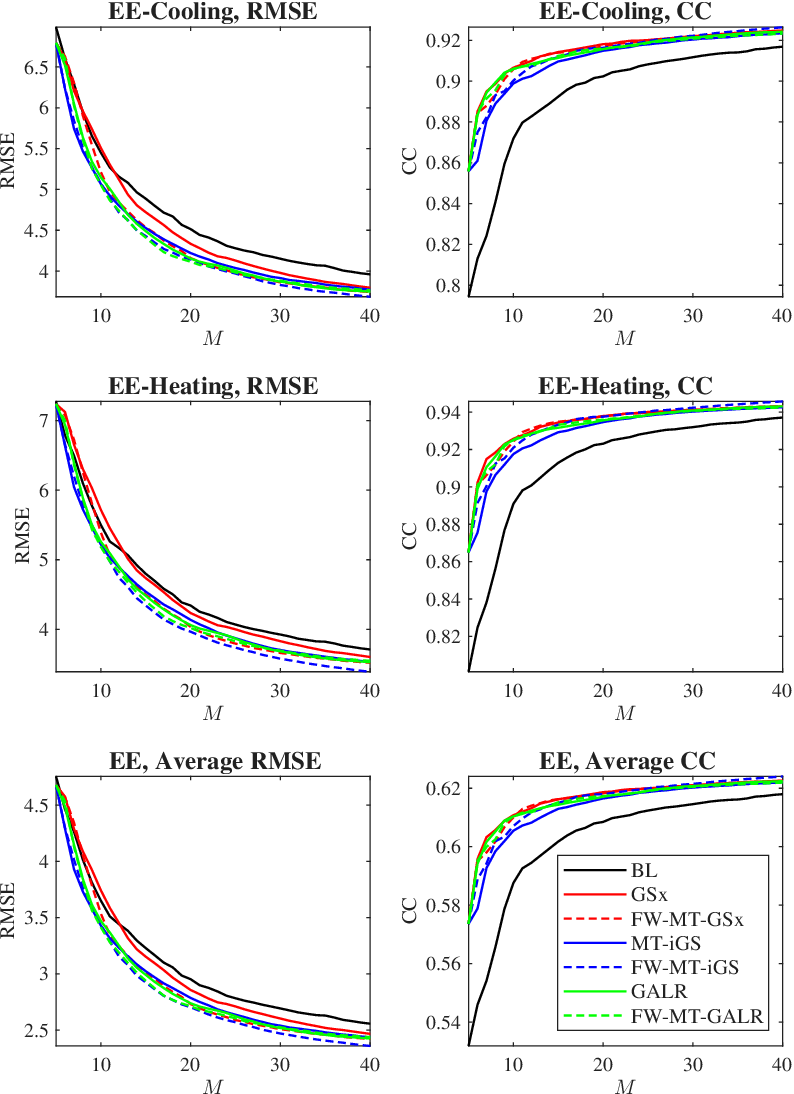}
\caption{Performance of the seven multi-task ALR algorithms on the EE dataset, averaged over 100 runs.} \label{fig:EE}
\end{figure}

Fig.~\ref{fig:EE} shows the RMSEs and CCs of the seven multi-task ALR algorithms on the EE dataset, averaged over 100 runs. Similar observations as in VAM can be made.

\subsection{Discussion}

The original multi-task ALR algorithms, \texttt{GSx}, \texttt{GALR} and \texttt{MT-iGS}, prioritize balance across different tasks, potentially overlooking samples that are particularly informative for individual tasks. They may be improved by weighting more important tasks more. For example, in (\ref{eq:fwmtdnmx}), $\left\|\mathbf{x}_i^{\mathrm{w}_t}-\mathbf{x}_j^{\mathrm{w}_t}\right\|^2$ instead of $\left\|\mathbf{x}_i^{\mathrm{w}_t}-\mathbf{x}_j^{\mathrm{w}_t}\right\|$ could be used to emphasize a particular task $t$.

However, no matter whether the multiple tasks have balanced or imbalanced importance, the effectiveness of feature weighting may not be affected. As an example,  compare Fig.~\ref{fig:results12}(g), where the task of EE-Cooling is considered without considering the task of EE-Heating at all (i.e., EE-Cooling has the maximum importance, whereas EE-Heating has the minimum importance), and Fig.~\ref{fig:EE}, where both tasks of EE-Cooling and EE-Heating are considered equally important. In both cases, feature weighted ALR algorithms always outperformed their unweighted counterparts.

\section{Conclusions and Future Research} \label{sect:conclusions}

In many real-world machine learning problems, acquiring a large amount of unlabeled data is relatively easy, but obtaining their labels is labor intensive, time consuming, and/or expensive. Active learning is a promising strategy to cope with these challenges. Particularly, pool-based sequential ALR optimally selects a small number of samples sequentially from a large pool of unlabeled samples to label, so that a more accurate regression model can be constructed under a given (small) labeling budget. Representativeness and diversity, which involve computing the distances among different samples, are important considerations in ALR. However, previous ALR approaches do not incorporate the importance of different features in inter-sample distance computation, resulting in inaccurate distances and hence sub-optimal sample selection.

This paper has proposed four feature weighted single-task ALR approaches and three feature weighted multi-task ALR approaches, where the ridge regression coefficients trained from a small amount of previously labeled samples are used to weight the corresponding features in inter-sample distance computation. Extensive experiments showed that this easy-to-implement enhancement consistently improved the performance of five existing ALR approaches, in both single-task and multi-task regression problems.

The following directions will be considered in our future research:
\begin{enumerate}
\item Theoretical justifications of the advantages of feature weighting in ALR. This paper has provided extensive experiment results to demonstrate the effectiveness and robustness of feature weighting, but in-depth theoretical justifications would make it more complete.
\item Incorporating feature weights in stream-based ALR~\cite{Cacciarelli2023,Cacciarelli2024,Horiguchi2024}, where unlabeled data arrive continuously, and we need to determine on-the-fly if a sample should be selected for labeling. The scenario is significantly different from pool-based ALR; however, as long as unlabeled sample selection involves computing the distances among samples, feature weighting should still be beneficial.
\item Incorporating feature weights in active learning for classification problems. As in ALR, inter-sample distance computation is also frequently used in active learning for classification problems to assess the representativeness and diversity of the unlabeled samples. So, the feature weighting strategy proposed in this paper may also be used to improve the performance of active learning for classification problems.
\item Considering sample weights in active learning approaches. Many active learning approaches, e.g., RD and FW-RD, use clustering to select the most representative and diverse samples, usually one sample from each cluster; however, different clusters have different sizes, and existing approaches do not take the cluster size into consideration. Weighting the selected samples by their corresponding cluster sizes may further improve the learning performance.
\end{enumerate}


\end{document}